\documentclass{article}

\usepackage[preprint]{template}

\usepackage{enumitem}
\usepackage[utf8]{inputenc} 
\usepackage[dvipsnames]{xcolor}

\usepackage[T1]{fontenc}    
\usepackage{hyperref}       
\usepackage{url}            
\usepackage{booktabs}       
\usepackage{amsfonts}       
\usepackage{nicefrac}       
\usepackage{microtype}      
\usepackage{tikz}
\usetikzlibrary{shapes,arrows,positioning,calc,fit}
\usetikzlibrary{decorations.pathmorphing}
\usetikzlibrary{arrows.meta,bending}
\usepackage{amsmath}
\usepackage{wrapfig}
\usepackage{algorithm}
\usepackage{algpseudocode}
\usepackage{dsfont}

\title{Model alignment using inter-modal bridges}

\author{Ali Gholamzadeh, \\
  MPI for Biological Cybernetics  \\
  \& University of Tübingen \\
   \texttt{ali.gholamzadeh@tue.mpg.de} \\
  \And
  Noor Sajid \\ 
  Kempner Institute, 
  Harvard University  \\
  \&  MPI for Biological Cybernetics\\
  \texttt{noorsajid@g.harvard.edu} \\
}

\begin{document}

\maketitle

\begin{abstract}
Foundation models have demonstrated remarkable performance across modalities such as language and vision. However, model reuse across distinct modalities (e.g., text and vision) remains limited due to the difficulty of aligning internal representations. Existing methods require extensive paired training data or are constrained to specific domains. We introduce a semi-supervised approach for model alignment via conditional flow matching. The conditional flow between latent spaces of different modalities (e.g., text-to-image or biological-to-artificial neuronal activity) can be learned in two settings: ($1$) solving a (balanced or unbalanced) optimal transport problem with an inter-space bridge cost, and ($2$) performing memory-efficient alignment using labelled exemplars. Despite being constrained by the original models' capacity, our method--under both settings--matches downstream task performance of end-to-end trained models on object recognition and image generation tasks across MNIST, ImageNet, and \cite{majaj2015simple} datasets, particularly when labelled training data is scarce ($<20\%$). Our method provides a data-efficient solution for inter-modal model alignment with minimal supervision.
\end{abstract}

\section{Introduction}\label{introduction}
\vspace{-0.8em}
Foundation models like GPT-X, DeepSeek, Gemini, Sora, and Dall-E have demonstrated remarkable performance across modalities such as text, image, and video~\citep{brown2020language,achiam2023gpt,team2023gemini}. While these large-scale models represent significant investments in computational resources and data curation~\citep{brown2020language}, their reuse across distinct data modalities, such as using a vision model on text data, remains limited by the fundamental challenge of aligning internal representations~\citep{imani2021representation,klebe2023gera,huh2024platonic}. Existing approaches for aligning models across modalities typically require extensive paired datasets to build correspondence~\citep{zhai2022lit}, yet such datasets are rarely available at scale~\citep{gadre2023datacomp}. Current alignment techniques are further constrained by their reliance on abundant paired data or their focus on specific domains~\citep{gadre2023datacomp}, limiting their broader applicability. Thus, developing alignment methods that can operate with minimal supervision while generalising across domains remains an open and critical challenge.

\begin{figure}[!t]
\centering
\begin{tikzpicture}[scale=0.8]
    \fill[red!10] (0,0) circle (1cm);
    \node at (0,0) {\(x_i^p\)};
    \draw[fill=red!10,  draw=red!10, line width=0pt] (-3.5,.5) -- (-3.5,-.5) -- (-2.5,-0.25) -- (-2.5,0.25) -- cycle;
    \node at (-3,0) {\(f_{\mathcal{X}}\)};
    \filldraw[red] (-0.7,-0.5) circle (1pt);
    \filldraw[red] (0.2,-0.5) circle (1pt);
    \filldraw[red] (-0.3,-0) circle (1pt);
    \filldraw[red] (0.21,0.8) circle (1pt);
    \filldraw[red] (0.3,0.7) circle (1pt);
    
    \fill[blue!10] (0,3) circle (1cm);
    \node at (-0.2,2.7) {\(y_j^p\)};
    \draw[fill=blue!10,  draw=blue!10, line width=0pt] (-3.5,3.5) -- (-3.5,2.5) -- (-2.5,2.75) -- (-2.5,3.25) -- cycle;
    \node at (-3,3) {\(f_{\mathcal{Y}}\)};
    \filldraw[blue] (0.4,3.7) circle (1pt);
    \filldraw[blue] (0.5,2.7) circle (1pt);
    \filldraw[blue] (0.3,3.1) circle (1pt);
    \filldraw[blue] (-0.5,2.8) circle (1pt);

    \fill[cyan!10] (4,2) circle (0.7cm);
    \node at (4,2) {\(z_i\)};
    \filldraw[cyan] (3.7,2.1) circle (1pt);
    \filldraw[cyan] (3.75,1.5) circle (1pt);
    \filldraw[cyan] (3.9,2.5) circle (1pt);
    \filldraw[cyan] (4.3,2.4) circle (1pt);
    
    \node at (-4,0.1) {\(D_{\mathcal{X}}\)};
    \node at (-4,3.1) {\(D_{\mathcal{Y}}\)};
    
    \draw[->] (-3.75, 0) -- (-3.5,0) ;
    \draw[->] (-3.75, 3) -- (-3.5,3);
    \draw[->] (-2.5,0) -- (-1.,0);
    \draw[->] (-2.5,3) -- (-1.,3);

    \draw[<->, dashed, black] (-0.3,-0) --
    node[midway, left, font=\small, black] {\((x_i^p, y_j^p) \in P\)} (-0.5,2.8);
    \draw[<->, dashed, black]  (0.5,2.7)-- (0.21,0.8);
    
    \draw[-{Triangle[width=24pt,length=10pt]}, line width=10pt, cyan!10] (3.5,2.4) to[bend right=20] node[midway, above, font=\small, black] {\(v_{t,\theta}\)} (1, 3);
    \draw[->, dotted, black] (3.7,2.1) to[bend right=20] (0.3,3.1);
    \draw[->, thick, red] (0.2,-.5) to[bend right=30] 
    node[midway, right, font=\small, black] {\(\pi^\star(\cdot|x_i)\)} (2.55,2.55);
    \draw[fill=red, draw=red, line width=2.5pt] (2.6,2.6) rectangle (2.5,3.1);

    \node[font=\small] at (0,-1.25) {source: \(\mu \in \mathcal{P}(\mathcal{X})\)};
    \node[font=\small] at (0, 4.2) {target: \(\nu \in \mathcal{P}(\mathcal{Y})\)};
    \node[font=\small] at (4.2, 3.1) {noise: \(\rho \in \mathcal{P}(\mathcal{Z})\)};
    
\end{tikzpicture}
\caption{Pictorial representation of our approach for aligning model space using inter-modal bridges. We consider two pre-trained models; \(f_\mathcal{X}\) and \(f_\mathcal{Y}\) and their corresponding datasets \(D_\mathcal{X}\) and \(D_\mathcal{Y}\). Next, we obtain source and target latent distributions \(\mu\) and \(\nu\) and compute the optimal coupling \(\pi^\star\) across these two distributions using paired samples \((x_i^p,y_j^p) \in P\) as inter-space bridges or using the paired sampled directly. Given this, we learn the velocity field \(v_{t,\theta}\) that morphs noise \(\rho\) conditioned on \(x_i\) in the source distribution to some target \(y_i\) using samples from the optimal coupling \(\pi^\star(\cdot \mid x_i)\).
\vspace{-1em}}
\label{fig:problem_formulation}
\end{figure}
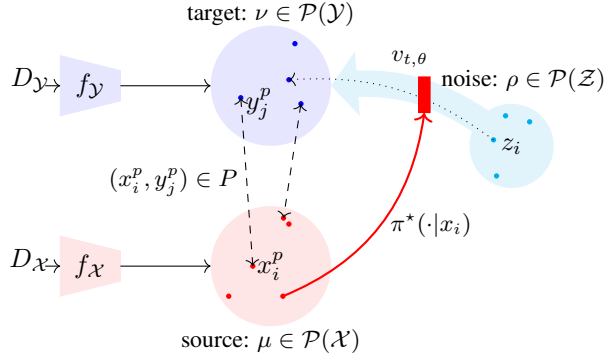

To address this, we propose \textit{model space alignment via inter-modal bridges} for integrating models across modalities with minimal supervision. Our approach centres on learning morph between latent spaces, where a noise distribution is mapped to the target space conditioned on source samples~\citep{klein2023generative}. These morphs are learned in a semi-supervised setting with access to a small set of paired samples between target and source distributions. We use these paired samples in two ways: through true alignment using the labelled pairs themselves, or by solving a balanced or unbalanced optimal transport (OT) problem~\citep{peyre2019computational} that uses our inter-modal bridge cost~(Sec.~\ref{sec:method-m3-bridge}) to compute optimal couplings between distributions. The inter-modal bridge cost captures similarities between latent spaces using intra-space distances and paired samples. 

Our contributions are as follows, we: $a)$ introduce an inter-modal bridge cost across distinct latent spaces using intra-space distances and paired inter-space samples (Sec.~\ref{sec:method-m3-bridge}), $b)$ show improved conditional flow matching using global OT alignment and true alignment compared to a local OT alignment baseline~\citep{klein2023generative} (Sec.~\ref{sec:results}), and $b)$ validate on downstream object recognition and image generation tasks using minimal paired samples (i.e., $<20\%$) (Sec.~\ref{sec:results}).    

\vspace{-0.5em}
\section{Preliminaries}\label{sec:prelim}
\vspace{-0.5em}
Our objective is to learn a transport function \(T: \mathcal{X} \rightarrow \mathcal{Y}\) that pushes forward a distribution \(\mu \in \mathcal{P}(\mathcal{X})\) to \(\nu \in \mathcal{P}(\mathcal{Y})\) (Fig.~\ref{fig:problem_formulation}). These distributions are induced by two datasets, \( D_\mathcal{X} \) and \( D_\mathcal{Y} \), associated with bounded sets \(\mathcal{X} \subset \mathbb{R}^p\) and \(\mathcal{Y} \subset \mathbb{R}^q\), referred to as the source and target domains, respectively. Given a feature extractor, we map raw data into a latent space where \(T\) operates: \(f_\mathcal{X}: D_\mathcal{X} \rightarrow \mathcal{X}\). 
\vspace{-0.9em}
\paragraph{Conditional Flow Matching (CFM)} is a simulation-free method that learns continuous maps between distributions by training a time-dependent velocity field to approximate the displacement between samples drawn from the source and target~\citep{lipman2022flow}. The velocity field, \(v_{t,\theta}(x): [0,1] \times \mathcal{X} \rightarrow \mathbb{R}^p,\) describes the instantaneous direction and speed of mass transport at each point along the path from the source to the target distribution. It is learnt by optimising:
\begin{equation}
    \mathcal{L}_{\text{CFM}}(\theta) = \mathbb{E}_{t, x_1 \sim p_1, x \sim p_t(x|x_1)} \| v_{t,\theta}(x) - u_t(x|x_1) \|^2~,
\end{equation}
where \(p_1(x)\) is the target distribution and \(u_t(x\mid x_1)\) is the conditional vector field\footnote{ A common choice for this is the linear, \(
u_t(x \mid x_1)
= \frac{x_1 - (1 - \sigma_{\min})x}
       {1 - (1 - \sigma_{\min})t}.
\)} that induces the probability path \(p_t(x\mid x_1)\) (see Appendix~\ref{appendix:flow_matching} for details).
Building on CFM, \citet{tong_conditional_2023} introduced OT-CFM to distil the discrete optimal transport map between two distributions in the same space.
\vspace{-0.9em}
\paragraph{Generative Entropic Neural Optimal Transport (\textcolor{blue}{U}-GENOT)} extends OT-CFM to the cross-domain setting~\citep{klein2023generative}.
To model this, it defines a conditional time-dependent velocity field \( v_{t,\theta}(\cdot \mid x): \mathcal{Y} \rightarrow \mathcal{Y} \). This specifies transport directions in the target space, conditioned on source points \(x \in \mathcal{X}\). The parameters \(\theta\) are optimised via:
\begin{equation}\label{eq:genot}
\begin{aligned}    
    \mathcal{L}_{\text{\textcolor{blue}{U}-GENOT}}(\theta) = \mathbb{E}_{t, z \sim \rho, (x,y)~\sim \pi^\star} \| v_{t,\theta}( ty + (1-t)z \mid x) - (y - z) \|^2  \\
    + \textcolor{blue}{\mathbb{E}_{x \sim \mu} \left[ (\eta - \eta_{\theta})(x)^2 \right]} +  \textcolor{blue}{\mathbb{E}_{y \sim \nu} \left[ (\xi - \xi_{\theta})(y)^2 \right]},
\end{aligned}
\end{equation}
where \(t \sim U[0,1]\), \(\pi^\star\) is the optimal coupling found using any discrete OT solver (Appendix~\ref{appendix:opt} for details), $\rho $ is the noise distribution \( \mathcal{N}(0,I_q) \in \mathcal{P}({\mathbb{R}^q})\), and \textcolor{blue}{$\eta_{\theta}$, $\xi_{\theta}$} are the non-negative neural reweighting functions (Appendix~\ref{appendix:unbalanced_theory} for details). \textcolor{blue}{U}-GENOT learns a separate flow for each point in the source distribution \( x \in \mathcal{X} \) and transforms a noise distribution \( \rho \) into a conditional coupling \( \pi^\star(\cdot\mid x)\) (defined using an appropriate OT solver; Appendix~\ref{appendix:opt}) for out-of-sample prediction.

\vspace{-0.5em}
\section{Learning inter-modal morphs using bridge cost}\label{sec:method}
\vspace{-0.5em}
In this section, we outline our approach for learning inter-modal morphs\footnote{
Here, we use "morph" as a general term for mapping a source distribution to a target distribution, while "conditional flow matching" refers to the specific implementation used to learn this mapping.}.  Specifically, we employ \textcolor{blue}{U}-GENOT~\citep{klein2023generative} to learn a transport function that maps between the latent distributions of two distinct domains (i.e., vision and text), leveraging features extracted from particular models and paired data points~(Sec~\ref{sec:method-problem}). This involves selecting the appropriate alignment strategy~(Sec~\ref{sec:method-m3-align}) to either find the optimal coupling (\(\pi^\star\): Eq.~\ref{eq:LEOT}-\ref{eq:fgw}) using the new bridge cost function~(Sec~\ref{sec:method-m3-bridge}) or using the labelled pairs directly, and an augmented neural architecture for learning the velocity field~(Eq.~\ref{eq:genot};  Appendix~\ref{appendix:architecture_adap}).  
\vspace{-0.5em}
\subsection{Problem setting}\label{sec:method-problem}
\vspace{-0.5em}
We aim to learn a transport function \(T: \mathcal{X} \rightarrow \mathcal{Y}\) mapping the latent distribution \(\mu \in \mathcal{P}(\mathcal{X})\) to \(\nu \in \mathcal{P}(\mathcal{Y})\) (Fig.~\ref{fig:problem_formulation}). We assume that if  $x_i \sim \mu $ and $y_i \sim \nu$ are the latent representations of the same object across distinct domains, the transport function should satisfy $T(x_i) \approx y_i$.
Given this, we assume access to \(n\) source domain points, \(X = [x_i]_{i=1}^n\), \(m\) target domain points, \(Y = [y_j]_{j=1}^m\), such that the set of labelled paired points \(\{x_i^p, y_i^p\} \in P\) have size \(|P| = l\) and \(0 < l \leq \max(n, m)\). These labelled paired points serve as an anchor for learning the inter-space mapping. This setting accommodates both a supervised setting where \(l = n = m\), i.e., all points are paired across domains, and a semi-supervised setting where \(l < \min(n, m)\), allowing us to use both labelled paired and unlabelled data points to construct the inter-space cost (see Sec.~\ref{sec:method-m3-bridge}). 
\vspace{-0.5em}
\subsection{Alignment strategy}\label{sec:method-m3-align}
\vspace{-0.5em}

\begin{figure}[!t]
\centering
\begin{tikzpicture}[scale=0.5]
    \begin{scope}[xshift=-7cm, yshift=0cm]
        \fill[blue!10] (0,0) circle (1.5);
        \fill[red!10] (3,0) circle (1.5);
        \node at (1.5,-2) {A. True};
        \draw [black, thick, <->] (0.5,0.5) -- (2.5,0.5);
        \draw [black, thick, <->] (0.7,-0.3) -- (2.3,-0.3);
        \draw [black, thick, <->] (-0.3,-0.7) -- (3.3,-0.7);
        \node at (0,0) {\textbf{X}};
        \node at (3,0) {\textbf{Y}};
    \end{scope}
    
    \begin{scope}[xshift=0cm, yshift=0cm]
        \fill[blue!10] (0,0) circle (1.5);
        \fill[red!10] (3,0) circle (1.5);
        \node at (1.5,-2) {B. Global};
        \draw [black, dashed, thick, <->] (0.2,0.7) -- (2.8,0.7);
        \draw [black, dashed, thick, <->] (0.3,-0.1) -- (2.7,-0.1);
        \draw [black, dashed, thick, <->]  (0.1,-0.9) -- (2.9,-0.9);
        \node at (0,0) {\textbf{X}};
        \node at (3,0) {\textbf{Y}};
    \end{scope}
    
    \begin{scope}[xshift=7cm, yshift=0cm]
        \fill[blue!10] (0,0) circle (1.5);
        \fill[red!10] (3,0) circle (1.5);
        \node at (1.5,-2) {C. Local};
        \draw [black, dashed, <->]  (0.4,0.2) -- (2.3,0.4);
        \draw [blue, dashed] (0.7,0) circle (0.5);
        \draw [red, dashed] (2.3,0.3) circle (0.5);
        \draw [black, dashed, <->]  (0.5,-0.3) -- (3.2,-0.8);
        \draw [black, dashed, <->]  (0.9,-0.15) -- (3,-0.4);
        \draw [red, dashed] (3,-0.6) circle (0.48);
        \draw [blue, dashed] (0.2,-1) circle (0.5);
        \draw [black, dashed, <->]  (0.2,-1) -- (3.,-0.8);
        \node at (0,0) {\textbf{X}};
        \node at (3,0) {\textbf{Y}};
    \end{scope}
\end{tikzpicture}
\caption{Pictorial representation of alignment methods. Here, --- represents true pairs, \(x_{ip}, y_{ip} \in P\), \textbf{\(-~-\)} the coupling given some inter-space cost \(C_{XY}\) and OT solver. \textcolor{red}{$\circ$}, \textcolor{blue}{$\circ$} represent sample batches.}
\label{fig::method_align}
\end{figure}
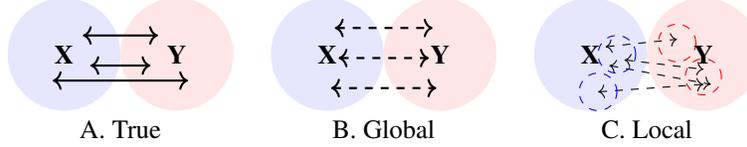
To compute the optimal coupling (\(\pi^\star\)) used to train the velocity field \(v_{t,\theta}\), we propose two distinct alignment strategies: true, and global (Fig. \ref{fig::method_align}). True alignment focuses on paired samples to construct a joint distribution between the source and target domains. Given a paired set \(P\), of size \(l\),
the coupling \(\pi^{\text{true}} \in \mathcal{P}(\mathcal{X} \times \mathcal{Y})\) is:
\vspace{-0.5em}
\begin{equation}\label{eq:true}
\pi^{\text{true}}(x_i, y_j) =
\begin{cases}
\frac{1}{l}, & \text{if } (x_i^p, y_j^p) \in P \\
0, & \text{otherwise}
\end{cases}
\end{equation}
This approach ensures direct alignment for paired data but does not fully use the information available in unpaired samples. Conversely, global alignment computes a full OT plan using the entire dataset. First, we construct an inter-space cost
matrix \(C_{XY}\) -- as defined in Sec~\ref{sec:method-m3-bridge} and Appendix~\ref{appendix:fused_cost} -- and then solve the OT problem:
$\pi^{\text{global}} = OT(X, Y, C_{XY})$, using the formulation defined in 
 Appendix~\ref{appendix:opt}. 
Global alignment provides an appropriate mapping but incurs a significant memory cost because the distance between each pair needs to be computed and stored. We compare with local alignment following~\citep{klein2023generative}. This solves the OT problem on subsets of the data at each iteration. For batches \(X^b\) and \(Y^b\) drawn from the source and target distributions, we compute:
$\pi^{\text{local}}_b = OT(X^b, Y^b, C_{XY}^b)~.$
This strategy improves scalability but can introduce misalignment due to batch approximations~\citep{fatras2021minibatch}.
This is because we are calculating a 'local' $\pi^{\text{local}}_b$ using a 'local' $C_{XY}^b$  at each iteration, and then sampling from this to learn the conditional flow. 
\vspace{-0.5em}
\subsection{Inter-modal Bridge cost}\label{sec:method-m3-bridge}
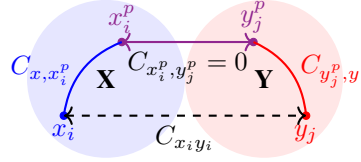
\begin{wrapfigure}{R}{0.54\textwidth}
\vspace{-10pt}
\centering
\begin{tikzpicture}[scale=0.7]
    \begin{scope}
        \fill[blue!10] (0,0) circle (1.5);
        \fill[red!10] (3,0) circle (1.5);
        \filldraw[Plum] (0.3,0.7) circle (2pt) node[above] {\(x_{i}^p\)};
        \filldraw[Plum] (2.8,0.7) circle (2pt) node[above] {\(y_{j}^p\)};
        \filldraw[blue] (-0.8,-0.7) circle (2pt) node[below] {\(x_{i}\)};
        \filldraw[red] (3.8,-0.7) circle (2pt) node[below] {\(y_{j}\)};
        \draw [blue, thick] (-0.8,-0.7) to[bend left] node[midway, left] {\(C_{x,x_{i}^p}\)} (0.3,0.7);
        \draw [red, thick] (2.8,0.7) to[bend left] node[midway, right] {\(C_{y_{j}^p,y}\)} (3.8,-0.7);
        \draw [Plum, thick, <->] (0.3,0.7) -- (2.8,0.7) node[midway, below, black] {\(C_{x_{i}^p,y_{j}^p} = 0\)};
        \draw [<->, thick, black, dashed] (-0.8,-0.7)  --(3.8,-0.7) node[midway, below] {\(C_{x_iy_i}\)};
        \node at (0,0) {\textbf{X}};
        \node at (3,0) {\textbf{Y}};
    \end{scope}
\end{tikzpicture}

\caption{Pictorial representation of bridge cost via \((x_{i}^p,y_{j}^p) \in P\) and \((x_{i},y_{j}) \not\in P\). }
\label{fig::method_fused}
\vspace{-25pt}
\end{wrapfigure}

\vspace{-0.5em}
We introduce a bridge cost, \(C^{\text{bridge}}_{XY}\), that uses paired samples to construct the inter-space cost, treating paired points as anchors linking the two spaces (Fig.~\ref{fig::method_fused}):
\begin{equation}\label{eq:bridge}
C^{\text{bridge}}_{XY}(x_i, y_j) =
\begin{cases}
0 & \text{if } (x_{i}^p, y_{j}^p) \in P \\
\min\limits_{(x_{i}^p, x_{j}^p) \in P} \left( C_{XX}(x_i, x_{i}^p) + C_{YY}(y_{j}^p, y_j) \right) & \text{otherwise}
\end{cases}
\end{equation}

where, \(C_{XX}\) and \(C_{YY}\) are the intra-space cost matrices and \(C^{\text{bridge}}_{XY}\) is determined as the minimum sum of these, using the paired samples as zero-cost links between the spaces. This represents the bridging cost between the paired elements. The intra-cost, \(C_{XX}\) and \(C_{YY}\) was calculated as the cosine distance between \(x_{i},x_j \in X\) and \(y_{i}, y_j \in Y\) \newline -- unless specified otherwise.
\vspace{-0.5em}
\subsection{Algorithm overview}
\vspace{-0.5em}
To learn the inter-modal morphs, for each modality we extract the features using some model, \(f_\mathcal{X}: D_\mathcal{X} \rightarrow \mathcal{X}\). The optimal coupling \(\pi^\star\) is calculated using one of the three proposed alignment strategies: true, global, or local (Sec.~\ref{sec:method-m3-align}). The training loop entails sampling from this coupling ($x_i, y_i \sim \pi^*$; $\pi^*$ is recalculated at each iteration for local alignment), the noise distribution ($z_i \sim \rho$), and optimising the velocity field to minimise the transport cost across some $T_{iter}$ iterations (using Eq.~\ref{eq:genot}). This continues until convergence or for $T_{iter}$ iterations. Once trained, the velocity field can be used for out-of-sample prediction from the source to the target space. See Appendix~\ref{appendix:architecture_algorithm} for pseudo-code.
\vspace{-0.5em}
\section{Dataset and Experimental Setup}\label{sec:experiment}
\vspace{-0.5em}
We evaluated our approach in two settings: alignment between image-text representations and between biological-artificial neural network representations. For image-text alignment, we used:
\vspace{-0.5em}
\begin{itemize}[leftmargin=*]
    \item \textbf{MNIST} contains \(60,000\) training and \(10,000\) test samples of handwritten digits across ten classes~\citep{lecun2010mnist}. For these experiments, we used \(50,000\) samples from the training set to train two variational auto-encoders (VAE)~\citep{Kingma2014vae}~(See Appendix~\ref{appendix:vae-pre-trained} for more details): \textit{VAE\(_\text{image}\)} for reconstructing images and \textit{VAE\(_\text{text}\)} for reconstructing the one-hot encoded labels. Afterwards, these trained networks were used as 'pre-trained' models and features were extracted for the \(10,000\) remaining training samples. We used these to train the model morph from the latent space of \textit{VAE\(_\text{text}\)} to the one of \textit{VAE\(_\text{image}\)} and vice-versa\footnote{Bi-directionality was modelled to evaluate how morphing between text-to-image vs image-to-text could differ and their influence on the downstream task performance.}. The test data were used to evaluate different morph formulations. 
    \item \textbf{ImageNet} contains approximately \(1.2\) million training samples across \(1,000\) classes~\citep{imagenet15russakovsky}. We used two pre-trained models: \textit{ViT-Base}~\citep{dosovitskiy2021imageworth16x16words}, a vision transformer, for image feature extraction, and \textit{MiniLM-L6}~\citep{wang2020minilm}, a pre-trained sentence encoder, for textual features. Image features were derived from the classification token of the final layer of ViT-Base, while the textual features were encoded in the format 'A photo of a \textit{class name}', following the protocol introduced in CLIP~\citep{radford2021clip}. We used $50\%$ train$/10\%$ validation split to train inter-modal morphs, and the remaining \(40\%\) for evaluation. 
    \vspace{-0.5em}
\end{itemize}

For (potentially noisier) alignment of biological-artificial neural representations, we used:
  \vspace{-0.5em}
\begin{itemize}[leftmargin=*]
    \item \textbf{\cite{majaj2015simple}} dataset that contains neural activity recordings from the visual area (V4) and the inferior temporal cortex (IT) of monkeys viewing distinct visual stimuli. The stimuli had eight categories, each containing $8$ core images, resulting in $64$ unique stimuli. Each stimulus was paired with $50$ randomly selected backgrounds, generating a final set of $3,200$ images. To reduce noise, neural activity for each unique stimulus was averaged across approximately $50$ presentations, with a minimum of $28$ repetitions per stimulus. We used randomly selected splits for train/validation/test datasets, $60\%/20\%/20\%$, that were consistently used for all analyses. Using this dataset, we considered how aligned the neural activity across a biological and artificial network could be when exposed to the same stimulus. To extract artificial neural representations, we used $6$ different pre-trained networks with CNN or transformer architecture (See Appendix~\ref{appendix:majaj}), which were among the best performing on Brain-Score~\citep{SchrimpfKubilius2018BrainScore}.
\vspace{-0.5em}
\end{itemize}

Using these datasets, we assessed the impact of alignment modalities (Sec~\ref{sec:results::image-text}-\ref{sec:results::bio-art}), model quality (Sec~\ref{results:quality_pre}), the inter-modal bridge cost (Sec~\ref{results:fuse}) (Sec~\ref{results:arch}) and velocity field architecture (Sec~\ref{results:arch}) for learning appropriate morphs. All reported experiments used \(5\) different random seeds and had a training budget of \(18\) hours on one $A100$ GPU. The models used for each experiment are presented in Appendix.~\ref{appendix:models}, with evaluation metrics incl. morph quality, and downstream task performance measures (Appendix.~\ref{appendix:metrics}).

\vspace{-0.5em}
\section{Results}\label{sec:results}
\vspace{-0.5em}
\begin{figure}[!t]
    \centering
    \includegraphics[width=\linewidth]{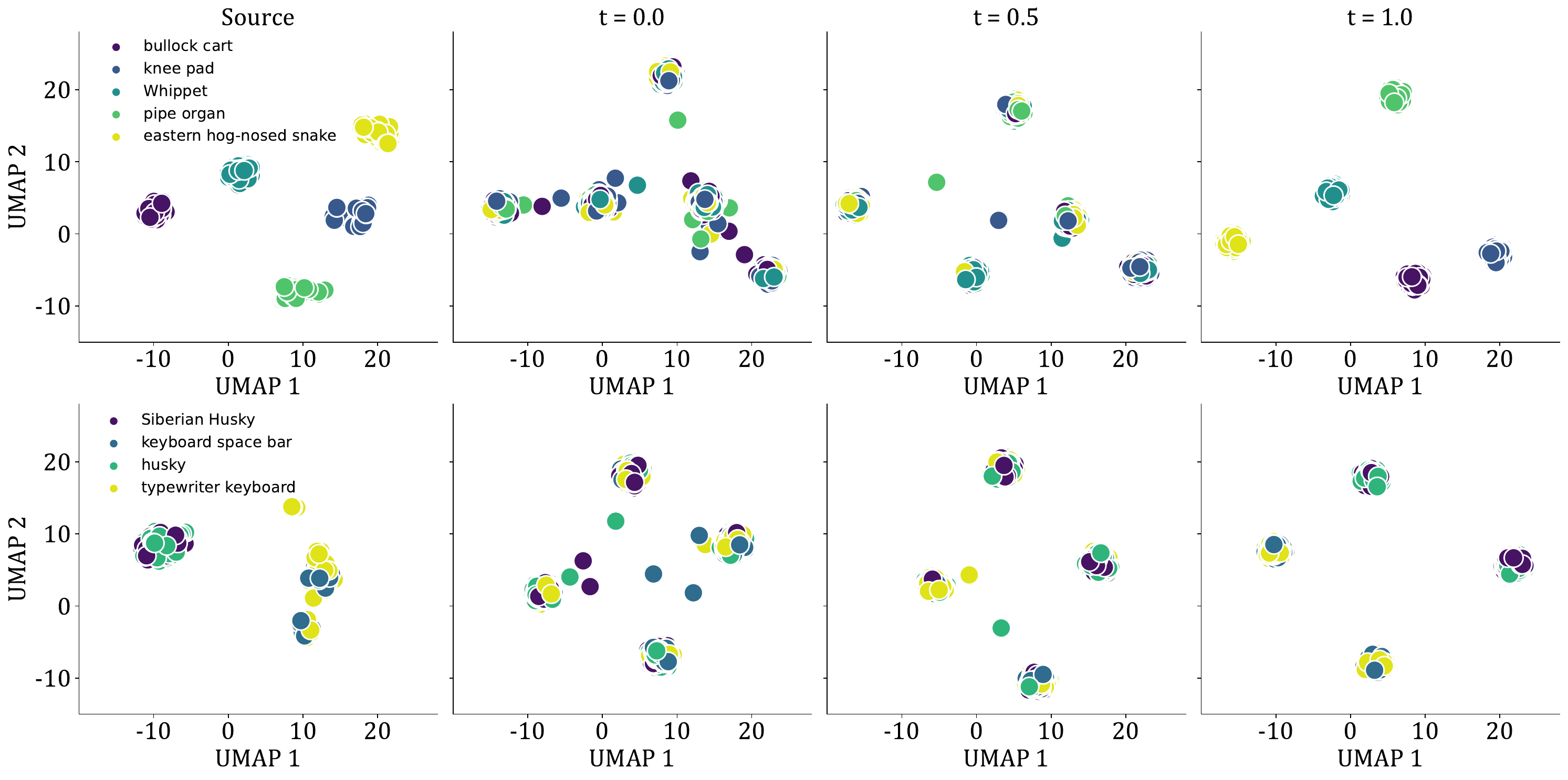}
    \caption{Noise distribution trajectory to the target latent space of a language model at \(t=(0,0.5,1)\) using \textit{true alignment} with \(1\%\) paired points. The latent source and target feature spaces for ImageNet are visualised using UMAP~\citep{mcinnes2018umap}. \textbf{Top:} Classes with minimal overlap in the image latent space. \textbf{Bottom:} Classes with high overlap in the image latent space.}
    \label{fig:qualitative}
\end{figure}
\subsection{Image-text alignment}\label{sec:results::image-text}
\vspace{-0.5em}
To align image-text representations, we used GENOT to learn conditional flow matching. In experiments with local and global alignment, the optimal coupling was computed using linear and FGW OT solvers.
\vspace{-0.5em}
\subsubsection{Image-to-text conditional flow matching}
\vspace{-0.5em}
We considered whether the degree of overlap in source feature space (i.e., images; Fig.\ref{fig:qualitative} source column) could influence the conditional flow matching to the target latent space (i.e., text; Fig.~\ref{fig:qualitative}, $t=1.0$ column) using UMAP projects of the feature spaces with $1\%$ paired points\footnote{We use UMAP to provide an intuitive low-dimensional representation, that captures simple correlations and local structure. Therefore, the UMAP visualisations should be interpreted as qualitative approximations rather than definitive measure of alignment quality.}. Using true alignment, we observed that when the source feature space (i.e., images) had minimal overlap, the resulting distribution showed a clear separation (Fig.~\ref{fig:qualitative}; top row, \(t=1.0\)). Conversely, when the feature spaces are similar (e.g., Husky and Siberian Husky), the resulting distribution mirrors the source, exhibiting substantial overlap (Fig.~\ref{fig:qualitative}; bottom row, \(t=1.0\)).
\vspace{-0.5em}
\subsubsection{Model 
quality}\label{results:quality_pre}
\vspace{-0.5em}
\begin{figure}[!t]
    \centering
    \includegraphics[width=0.8\linewidth]{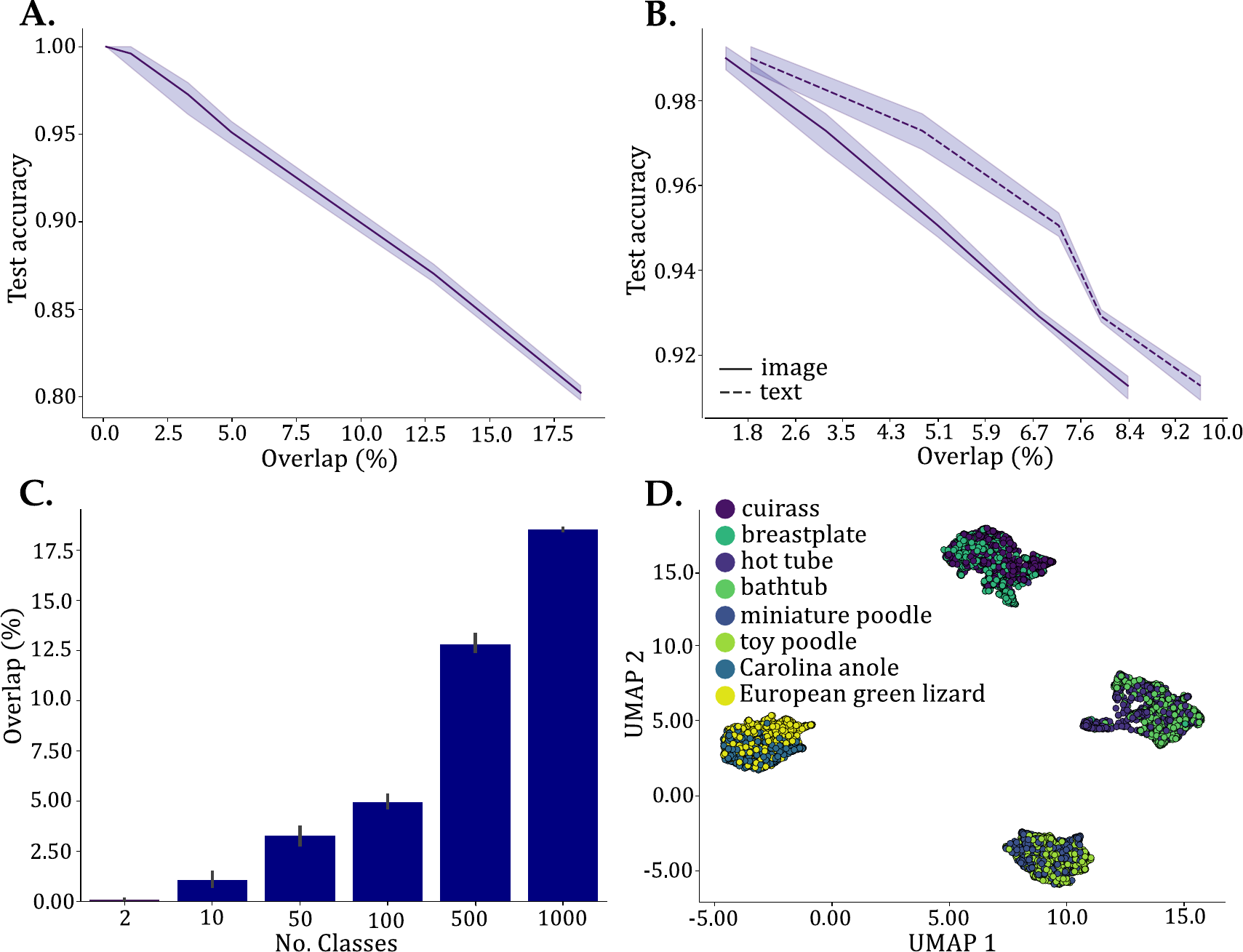}
    \caption{
   Latent space overlaps on conditional flow matching from image-to-text domain, using true alignment in a fully supervised setting. 
    \textbf{A)} MNIST experiment. 
    \textbf{B)} ImageNet experiment. 
    \textbf{C)} Relationship between the number of classes and the degree of overlap in the latent space for the ImageNet dataset. 
    \textbf{D)} UMAP visualisations of the classes with the highest overlap in the latent space of the ViT-B model.
    }
    \label{fig:quality_base}
\end{figure}
Building on this, we quantified how the quality of the pre-trained model influenced the performance of the learned flow from image to text. We measured quality in terms of feature space overlap, reflecting how well-disentangled the encoded space was, using varying numbers of randomly selected classes from each dataset (Appendix~\ref{appendix:overlap}). For both datasets, we observed a decline in performance (refer to Appendix~\ref{appendix:imagenet-acc-metric} for how accuracy was computed) as the feature space overlap increased (Fig.\ref{fig:quality_base}.A-B). The degree of overlap was directly correlated with the number of classes used (Fig.~\ref{fig:quality_base}.C). Fig.~\ref{fig:quality_base}.D presents a UMAP projection of the ViT-B feature space, highlighting several classes with significant overlap in the ImageNet experiments. This high overlap may result from the training capacity of the base ViT model~\citep{tsipras2020imagenet} or potential mislabelling within the original dataset~\citep{beyer2020we}. These results suggest that the latent spaces' degree of disentanglement and overall quality play a critical role in shaping the learned flow.
\vspace{-0.5em}
\subsubsection{Inter-space bridge 
cost}\label{results:fuse}
\vspace{-0.4em}
\begin{figure}[!b]
    \centering
    \includegraphics[width=0.8\linewidth]{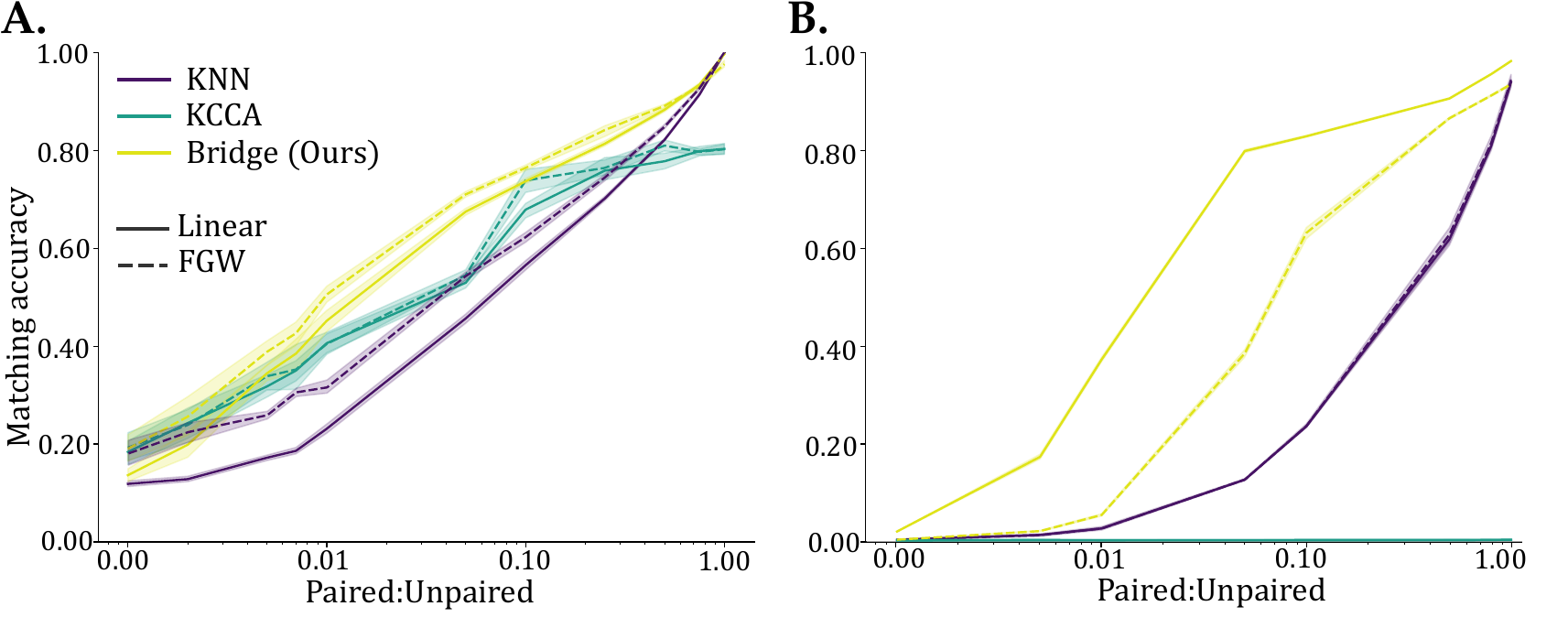}
    \caption{
    Matching accuracy across the inter-space costs for linear and FGW solvers. The optimised value of $\alpha^\star$ was used for FGW solvers. Matching accuracy was calculated by sampling from the optimal coupling \(\pi^\star\) and averaging the number of correct matches. \textbf{A)} MNIST experiment. $\alpha^\star$: KNN: $0.25$, KCCA: $0.5$, bridge: $0.5$. \textbf{B)} ImageNet experiment. $\alpha^\star$: KNN: $0.25$, KCCA: $0.5$,  bridge: $0.25$.}
    \label{fig:fuse}
\end{figure}   
\vspace{-0.4em}
We examined the effectiveness of the bridge cost function in learning the optimal coupling by evaluating the matching accuracy (i.e., the average number of correct responses based on the ground truth labels) for the Linear and FGW OT solvers. We compared the performance against KNN and KCCA cost functions (Appendix~\ref{appendix:fused_cost}), using large sample sizes (\(100,000\) for ImageNet and \(10,000\) for MNIST). Our results show that the bridge cost consistently outperformed the other cost functions across both datasets and various paired point configurations (Fig.~\ref{fig:fuse}). Additionally, we observe a clear association between the number of paired points and improvements in coupling quality and matching accuracy. Next, we evaluated the performance of discrete solvers. We find that FGW (Eq.\ref{eq:fgw}), which considers both intra- and inter-space costs using optimised \(\alpha^\star\)( Appendix.~\ref{appendix:alpha_opt}), performed the best for MNIST and linear OT solver (Eq.\ref{eq:LEOT}) for ImageNet. Based on these results, we employed the bridge cost and the best-performing OT solvers for all subsequent experiments.
\vspace{-0.5em}
\subsubsection{Velocity field network architecture}\label{results:arch}
\vspace{-0.1em}
\begin{figure}[!t]
    \centering
    \includegraphics[width=0.8\linewidth]{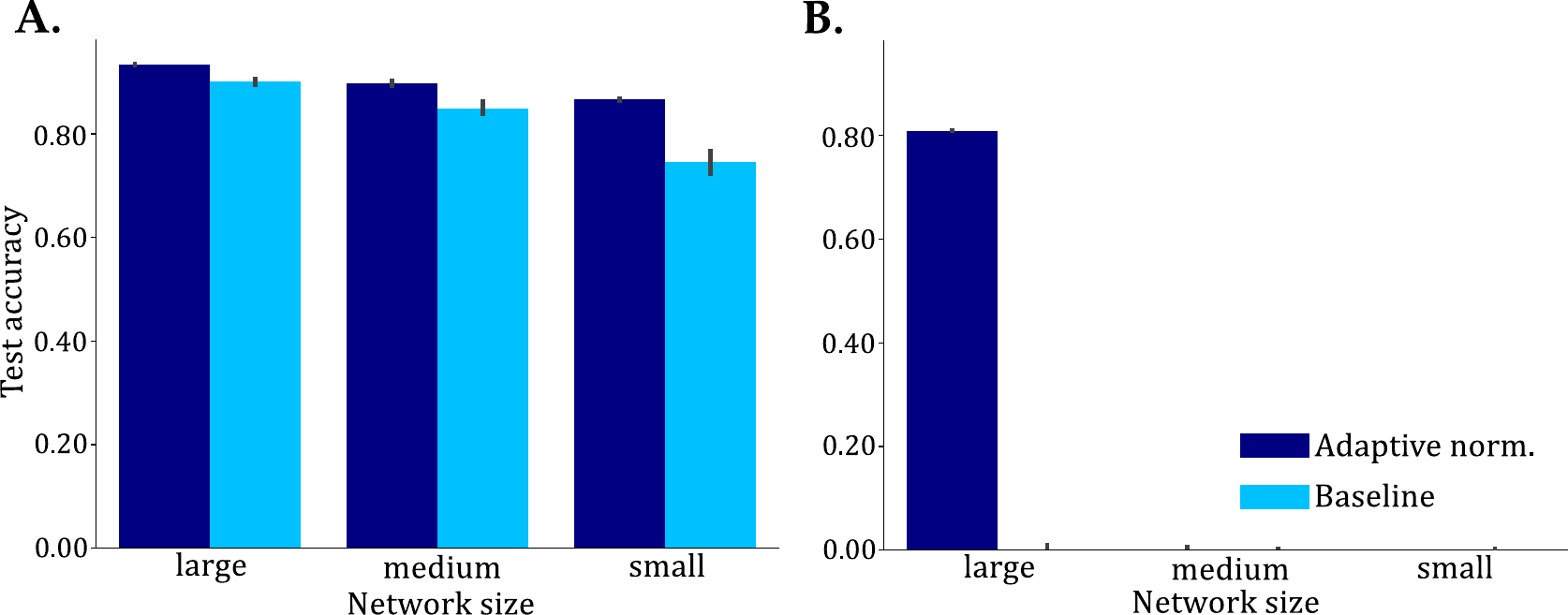}
    \caption{
    Velocity Field \(v_\theta\) architecture benchmark. 
    Comparison of the baseline feed-forward architecture and the adaptive normalisation architecture (with blocks) for image-to-text tasks in a fully supervised setting, using true alignment, across various sizes. Here, 
    \textbf{A}) MNIST experiments, and 
    \textbf{B}) ImageNet experiments. 
    \vspace{-0.5em}
    }
    \label{fig:architecture_size}
\end{figure}
\vspace{-0.5em}
We evaluated two distinct architectures -- \cite{klein2023generative} baseline and neural network with adaptive normalisation -- for parameterising the conditional velocity field network \( v_{t,\theta} \) (Appendix~\ref{appendix:architecture_adap}). The baseline follows the design of GENOT~\citep{klein2023generative}, utilising a neural network that takes time, source, and latent noise as inputs. Each input vector is embedded independently using a multi-layer perceptron (MLP) block before concatenation. The second architecture, inspired by Diffusion Transformers (DiT)~\citep{peebles2023dit}, integrates adaptive layer norm~\citep{perez2018film} (adaLN) blocks (see Appendix~\ref{appendix:architecture_adap}). Here, input latent noise is normalised in each block, conditioned on time and source data. We tested three size variants of each architecture. Our results show that the adaLN-based architecture consistently outperformed its counterpart (Fig.~\ref{fig:architecture_size}). For the MNIST dataset (Fig.\ref{fig:architecture_size}.A), all adaLN architectures outperformed the baseline architectures with similar parameter counts. For the ImageNet dataset, only the larger adaLN architecture successfully learned the mapping (Fig.\ref{fig:architecture_size}.B). Furthermore, in terms of sample efficiency, the larger adaLN networks achieved the target accuracy significantly faster -- potentially due to dynamic activation normalisation ( Appendix~\ref{appendix:architecture}). Given this, the velocity field in all remaining experiments was parameterised using the adaLN-Large architecture.

\begin{wrapfigure}{R}{0.45\textwidth}
    \centering
    \includegraphics[width=0.85\linewidth]{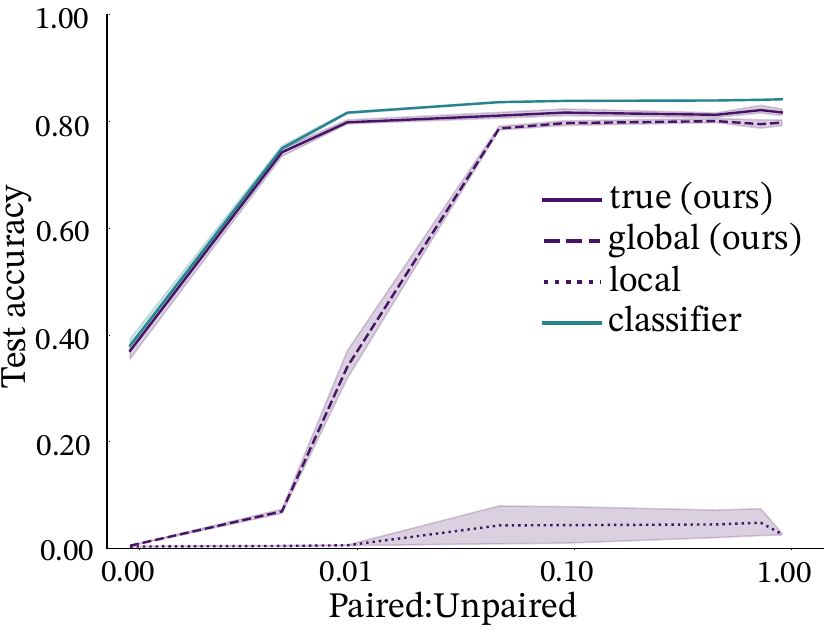}
    \caption{Image-to-text test accuracy for ImageNet.\vspace{-0.9em}
    }
    \label{fig:imagenet_results}
\end{wrapfigure}
\vspace{-0.5em}

\subsubsection{Downstream task performance}\label{results:downstream}
\vspace{-0.5em}
For the ImageNet dataset, we evaluated the accuracy of the image-to-text feature space using varying numbers of paired samples 
(Fig.~\ref{fig:imagenet_results}). We compared our approach against the classification head on ViT reported as $83.97$ in~\cite{dosovitskiy2021imageworth16x16words}. This represents the upper bound for the model's performance potential. Under the given training time constraints (i.e., \(18\) hours), local alignment failed to converge and exhibited poor performance, since it requires computing the inter-space cost and optimal coupling at each iteration~(see Appendix~\ref{appendix:time_complexity}). Similarly, the global solver underperformed in settings with very few paired samples, likely due to misalignment issues stemming from the discrete solver in these regimes (Fig.~\ref{fig:quality_base}). However, as the number of paired samples increased to \(\approx 10\%\), performance improved to a level comparable to the classifier.

\begin{figure}[!t]
    \centering
    \includegraphics[width=0.8\linewidth]{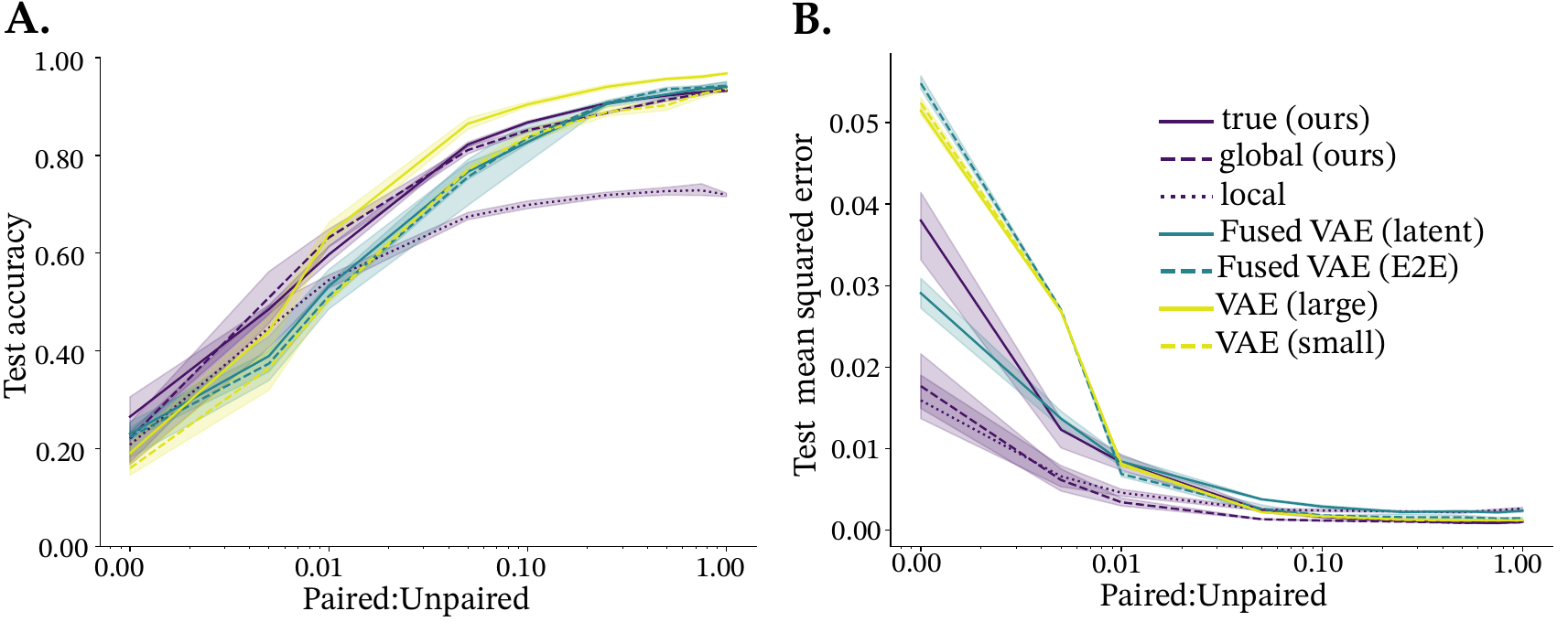}
    \caption{Downstream task performance across different alignment strategies using bridge cost with FGW solver, and the baseline models for MNIST. \textbf{A)} Test accuracy for text reconstruction from image input. \textbf{B)} Mean squared error between text-image constructions and true images pixel-wise 2D distributions for each class using test data.
    \vspace{-0.5em}}
    \label{fig:mnist_results}
\end{figure}

Finally, we compared the morph results across different numbers of paired points for the MNIST dataset. The image-to-text transformation revealed that global and true alignment performed on par with the baseline methods~(Fig.~\ref{fig:mnist_results}.A). Importantly, with a small number of paired samples (i.e., \(<10\%\)), our method outperformed these baselines. However, local alignment had significantly worse performance; due to misalignment and flow misguidance--also noted in~\citep{fatras2019learning}. This effect can be mitigated by increasing the batch size~\citep{klein2023generative}. For the text-to-image transformation (Fig.~\ref{fig:mnist_results}.B), we generated images for each class from the corresponding labels. After morphing from \textit{VAE\(_\text{text}\)} to the \textit{VAE\(_\text{image}\)} latent space, we reconstructed the images and evaluated them using mean squared error (Appendix~\ref{appendix:mnist}). In limited paired samples settings, the local and global alignment methods outperformed all other approaches.
\vspace{-0.5em}

\subsection{Biological-artificial neural representation alignment}\label{sec:results::bio-art}
\vspace{-0.5em}
Our initial evaluation highlighted that unbalanced OT is more applicable for noisy data settings (Appendix.~\ref{appendix:unbalanced_exp}). Therefore, we used: $1)$ \textcolor{blue}{U}-GENOT (Eq.~\ref{eq:genot}) to learn the conditional flow matching, $2)$ \textcolor{blue}{U}-EOT solver\footnote{We used linear since it performed better than \textcolor{blue}{U}-FGW, and following~\cite{klein2023generative} optimise the degree of mass conservation between source and target distributions via $\tau_\alpha$ and $\tau_\beta$ (Eq.~\ref{eq:genot}; Appendix~\ref{appendix:unbalanced}).} to compute the optimal coupling for global alignment and compared it to true alignment, and $3)$ used Pearson correlation to calculate the intra-space cost (Appendix.~\ref{appendix:inter-space_majaj}).

\begin{figure}[h]
    \centering \includegraphics[width=1\linewidth]{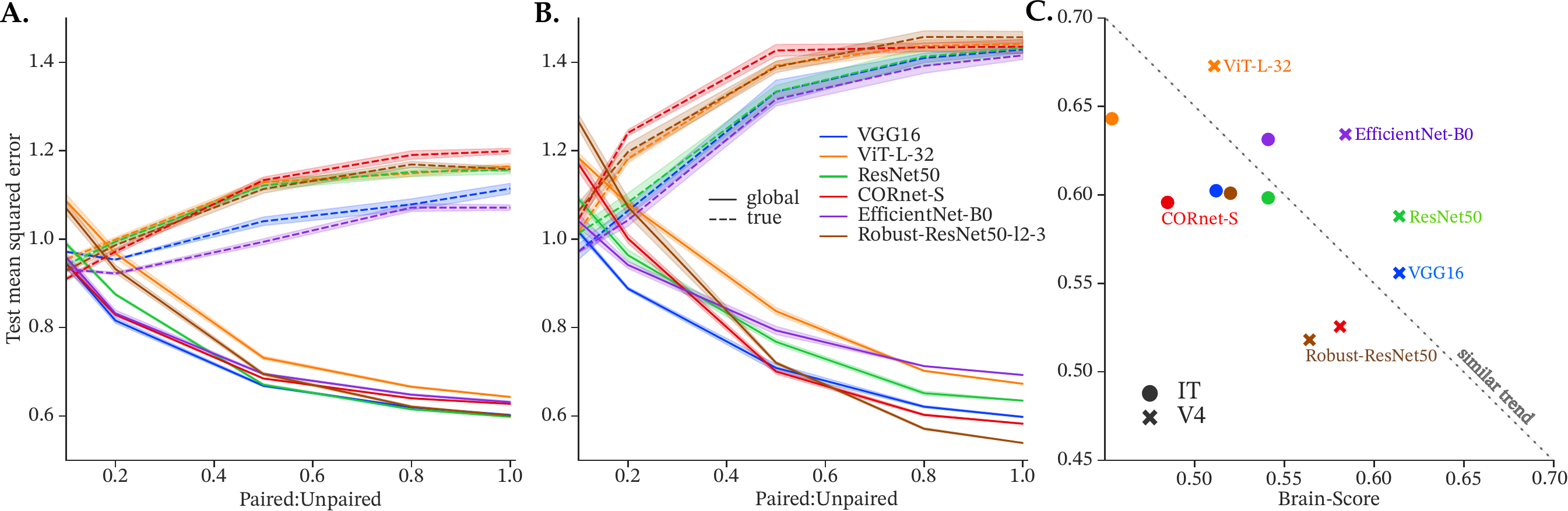}
    \caption{Test mean squared error of the learned conditional flow matching from representations of the best-performing layer of different neural networks to neural recordings in \textbf{A)} IT and \textbf{B)} V4, using true and global alignment. \textbf{C)} Comparison with Brain-Score in the fully supervised setting using global alignment.}
    \label{fig:majaj_pe}
\end{figure}
\subsubsection{Model-to-neural activity conditional flow matching}
\vspace{-0.5em}
To assess the quality of the learnt conditional flow matching across the different models (Table~\ref{tab:arch-majaj}), we compared the test mean squared error (MSE) under different alignment strategies, true and global, while varying the proportion of paired data using the best-performing layers. For true alignment, increasing the proportion of paired data led to over-fitting, as reflected in a consistent increase in MSE when morphing between the artificial models and neural activity (Fig.~\ref{fig:majaj_pe}). Conversely, for the global alignment, an increase in the proportion of paired data corresponds to a decrease in MSE (Fig.~\ref{fig:majaj_pe}). As expected, mappings from the later layers to the IT region have a lower MSE, while the early or middle layers to the V4 have a lower MSE (Fig.~\ref{fig:majaj_pe_layers}). These results are not consistent with Brain-Score benchmark (Fig.~\ref{fig:majaj_pe}C) despite the best performing model (ResNet50) for IT being the same for both. 
\vspace{-0.5em}
\begin{figure}[!t]
    \centering \includegraphics[width=1\linewidth]{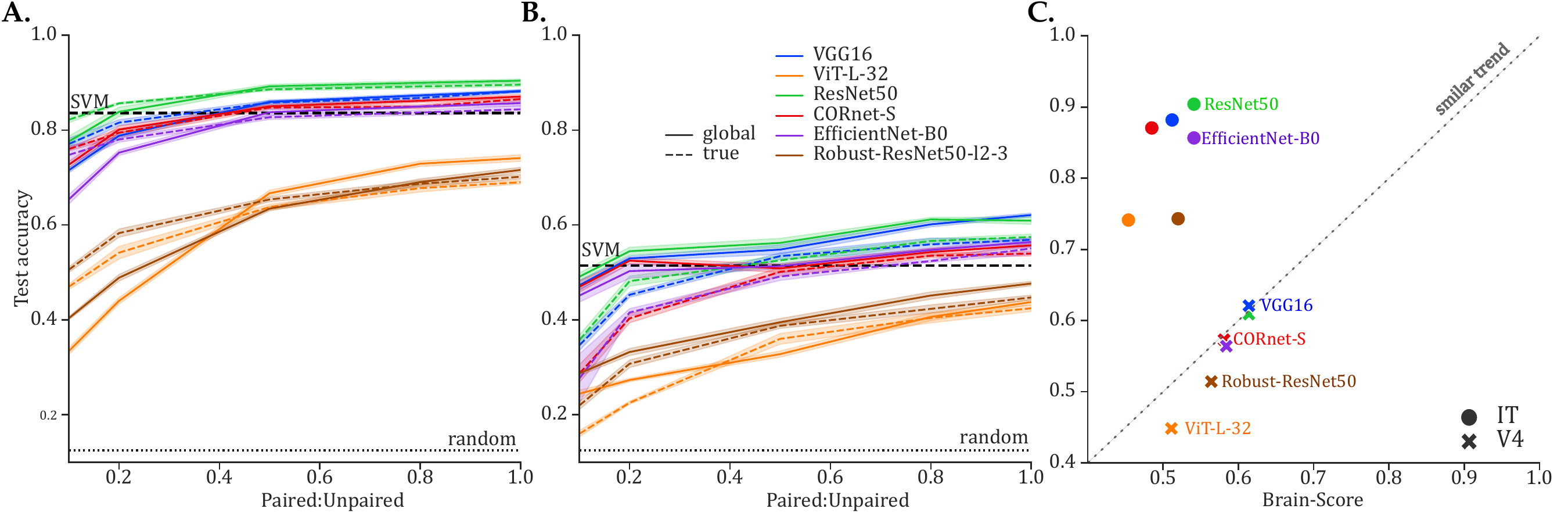}
    \caption{Downstream task performance for image category classification using the learned conditional flow maps from the best-performing layers of different artificial networks, mapped to neural recordings from \textbf{A)} IT and \textbf{B)} V4 regions. Neural activity was decoded using an SVM following the approach of \citet{majaj2015simple}; the SVM was trained to predict image categories from the neural representations. The black dotted line indicates random chance accuracy, while the black dashed line represents the SVM's accuracy on the actual brain activity test set. \textbf{C)} Comparison of test accuracy with Brain-Score in the fully supervised setting for the best-performing layers.}
    \label{fig:majaj_class}
\end{figure}
\subsubsection{Downstream task performance}
\vspace{-0.5em}
We evaluated category classification accuracy across different images using varying numbers of paired samples (Fig.~\ref{fig:majaj_class}) for the best-performing layers of different artificial neural networks (Appendix~\ref{appendix:majaj_svm} for evaluation details). 
Downstream task performance increased with the level of supervision for both true and global alignment. For IT and V4, deeper layers performed better on the downstream task (Fig.~\ref{fig:majaj_acc_layers}). This aligns with prior findings that later layers in an artificial neural network encode higher-order semantic features, making them more relevant for predicting activity in higher-order brain areas~\citep{yamins2014performance}. These results are fairly consistent with Brain-Score (Fig.~\ref{fig:majaj_class}C). Importantly, a biologically-inspired network (CORNet-S) has competitive accuracy using our alignment setting compared to other networks (e.g., Robust-ResNet50). 
\vspace{-0.5em} 

\section{Conclusion}\label{sec:conclusion}
\vspace{-0.9em}
We investigated inter-modal model alignment across text-image and biological-artificial neural representations. For this, we introduced an inter-modal bridge cost for fusing feature spaces. Our results show that this enables effective alignment between distributions from separate modalities, even with limited paired samples between source and target spaces.
We found that global alignment (using samples from the computed optimal coupling) achieves competitive downstream performance while avoiding overfitting in noisy settings compared to true alignment (which only used labelled pairs). These findings emphasise two key factors for morphing quality: intra-space separation within feature spaces and inter-space alignment between them. However, the effectiveness of our method may be limited by the quality of pre-trained feature extractors and the availability of paired samples. Future work should focus on developing more disentangled representations to improve model reusability across modalities. Separately, our future work will extend our alignment approach to the scenarios where the assumption about paired data points is relaxed. 

\section*{Acknowledgements}

The authors thank Peter Dayan for his valuable feedback  on the manuscript. This work was supported by the Max Planck Society. 



\bibliographystyle{plainnat}
\bibliography{main}


\newpage
\appendix
\section*{Supplementary Materials: Model alignment using inter-modal bridges}

\section{Societal impact}\label{appendix:societal}
Our method improves model reuse across modalities with minimal supervision, enabling applications in low-resource settings. However, it may also pose risks, such as enabling surveillance via cross-modal linking of personal data or amplifying biases when aligning poorly disentangled representations. Ethical safeguards are advised.

\section{Primer on Optimal transport}\label{appendix:opt}
\paragraph{Background} Optimal Transport (OT) was introduced by \citep{monge1781} as a way of transferring dirt from one place to another by minimising the transport cost between the source and target distributions. Given two probability measures \( \mu\in \mathcal{P}(\mathcal{X})\)and \(\ \nu \in \mathcal{P}(\mathcal{Y})\), and a cost function \(c(x, y)\) that quantifies the distance between pairs \((x, y)\) where \(x \in X\) and \(y \in Y\), the objective is to find a push-forward map \(T\) that minimises the total cost of transporting mass from \(\mu\) to \(\nu\). This problem can be mathematically formulated as finding the optimal map \(T^\star\) that solves:
\begin{equation}    
T^\star := \arg \inf_{T_\#\mu = \nu} \int_X c(x, T(x)) d\mu(x)~,
\end{equation}
subject to the constraint that the push-forward of \(\mu
\) under \(T\) equals \(\nu\). However, solving the Monge problem is challenging, and the map \(T^\star\) may not be unique or even exist in some cases. Kantorovich introduced a relaxation of the original Monge problem~\citep{kantorovich1942transfer} i.e., instead of seeking a deterministic mapping between two distributions, Kantorovich proposed finding a probabilistic mapping \(\pi\), known as a coupling, which is a joint probability distribution over \(\mathcal{X} \times \mathcal{Y}\), and its marginal distribution is denoted as \(\pi_{\mathcal{X}}(x) = \int_{\mathcal{Y}} \pi(x, y) \, dy\). Then, the Kantorovich problem is defined as:
\begin{equation} \label{eq:kantorovich}
    \pi^\star := \arg \inf_{\pi\in \Pi(\mu, \nu)}\int_{\mathcal{X} \times \mathcal{Y}} c(x, y) \pi(x, y)\mathrm{d}x\mathrm{d}y.
\end{equation}

For computational efficiency, the entropy-regularised version of this problem is usually considered in OT formulations. 

\paragraph{(\textcolor{blue}{Unbalanced}) Linear entropic OT} Given some cost function \(c: \mathcal{X} \times \mathcal{Y} \rightarrow \mathbb{R}\), 
the (\textcolor{blue}{unbalanced}) linear entropic OT for \(\mu \in \mathcal{P(\mathcal{X})}\) and  \(\nu \in \mathcal{P(\mathcal{Y})}\) is:
\begin{align}\label{eq:LEOT}
\vspace{-0.5em}
\pi^\star := \arg \inf_{\pi\in \mathcal{P}(\mathcal{X} \times \mathcal{Y})}\int_{\mathcal{X} \times \mathcal{Y}} c(x, y) \pi(x, y)\mathrm{d}x\mathrm{d}y - \epsilon H(\pi) 
+ \textcolor{blue}{\lambda_{\mathcal{X}}\text{KL}(\pi_{\mathcal{X}} \parallel \mu)} 
+ \textcolor{blue}{\lambda_{\mathcal{Y}}\text{KL}(\pi_{\mathcal{Y}} \parallel \nu)}~,
\vspace{-0.9em}
\end{align}
where \(\epsilon \geq 0\) is a hyperparameter controlling the trade-off between minimising the transport cost and the smoothness of the solution, \(\textcolor{blue}{\lambda_{i}\text{s}}\) the unbalanced weighting parameters\footnote{In our experiments, instead of directly setting the \textcolor{blue}{\(\lambda_i\)}'s, we use an alternative parameter \textcolor{blue}{\(\tau\)} (see Appendix~\ref{appendix:unbalanced_theory})}. The entropy for \(\pi \in \mathcal{P}(\mathcal{X} \times \mathcal{Y})\) is given by \(H(\pi) = - \int_{\mathcal{X} \times \mathcal{Y}} \pi(x, y) \log(\pi(x, y)) \, d(x,y).\) For \(p,q \in \mathcal{P}(X)\), \(\text{KL}(p \parallel q) = \int_X p(x) \log \frac{p(x)}{q(x)} \, dx\) denotes the Kullback--Leibler divergence between these two distributions.

For the discrete setting, the Sinkhorn algorithm~\citep{cuturi2013sinkhorn} solves the linear entropic OT problem by iteratively updating the coupling to minimise the regularised cost while satisfying marginal constraints. In the \textcolor{blue}{unbalanced} setting a variation of the Sinkhorn algorithm can be used~\citep{frogner2015learning, sejourne2023unbalanced}. Note, that adding regularisation into the Kantorovich formulation (Eq.~\ref{eq:kantorovich}) improves computational efficiency and stability by making the problem more tractable, often leading to smoother and more robust solutions. Additionally, regularisation can improve convergence properties and ensure the existence of unique solutions\citep{peyre2019computational}.

\paragraph{(\textcolor{blue}{Unbalanced}) Quadratic entropic OT}
Given two intra-space cost functions \(c_\mathcal{X} : \mathcal{X} \times \mathcal{X} \rightarrow \mathbb{R}\) and \(c_\mathcal{Y} : \mathcal{Y} \times \mathcal{Y} \rightarrow \mathbb{R}\), this method extends the linear OT problem to distinct spaces by learning a coupling that encourages the matches of elements close in one probability distribution to be close in the other distribution as well~\citep{vayer2020contribution,sejourne2023unbalanced}: 
\begin{equation}\label{eq:gw}
  \begin{aligned}
 \pi^\star := 
 &\arg \inf_{\pi \in \mathcal{P}(\mathcal{X} \times \mathcal{Y})} \int_{{(\mathcal{X} \times \mathcal{Y})}^2} \left| c_\mathcal{X}(x, x') - c_\mathcal{Y}(y, y') \right|^2 \, d\pi(x, y) \, d\pi(x', y') - \epsilon H(\pi) \\
 &+ \textcolor{blue}{\lambda_{\mathcal{X}}\text{KL}^\otimes
(\pi_{\mathcal{X}} \parallel \mu)} 
+ \textcolor{blue}{\lambda_{\mathcal{Y}}\text{KL}^\otimes(\pi_{\mathcal{Y}} \parallel \nu)}~,
\end{aligned}  
\end{equation}
where tensorised \(\text{KL}^\otimes(p \parallel q) = \text{KL}(p \otimes p \parallel q \otimes q)\). When \(\epsilon=0\), this reduces to the Gromov-Wasserstein~(GW)~\citep{memoli2011gromov} setting where the two spaces have different support\footnote{The Quadratic entropic OT is a non-convex optimization problem and for the discrete case has a time complexity of \(O(n^2)\) or \(O(n^3)\) \citealp{scetbon2022linear}}.

\paragraph{(\textcolor{blue}{Unbalanced}) Fused Gromov-Wasserstein (\textcolor{blue}{U}-FGW)} Given two partially comparable spaces, \textcolor{blue}{U}-FGW extends \textcolor{blue}{U}-GW by combining both the intra-space structural dissimilarity with an inter-space feature discrepancy~\citep{titouan2019optimal} and can be formalised as:
\begin{equation}\label{eq:fgw}
\begin{aligned}
\pi^\star :=
 & \arg \inf_{\pi \in \mathcal{P}(\mathcal{X} \times \mathcal{Y})} \int_{{(\mathcal{X} \times \mathcal{Y})}^2} \left( \alpha|c_\mathcal{X} - c_\mathcal{Y}|^2 + (1-\alpha) c_{\mathcal{X}\mathcal{Y}}^2\right) d\pi(x, y) \, d\pi(x', y') - \epsilon H(\pi) \\
 &+ \textcolor{blue}{\lambda_{\mathcal{X}}\text{KL}^\otimes
(\pi_{\mathcal{X}} \parallel \mu)} 
+ \textcolor{blue}{\lambda_{\mathcal{Y}}\text{KL}^\otimes(\pi_{\mathcal{Y}} \parallel \nu)}~,
\end{aligned}  
\end{equation}
where 
\(c_\mathcal{X} : \mathcal{X} \times \mathcal{X} \rightarrow \mathbb{R}\) and \(c_\mathcal{Y} : \mathcal{Y} \times \mathcal{Y} \rightarrow \mathbb{R}\) are the intra-cost functions: \(c_{\mathcal{X}\mathcal{Y}}: \mathcal{X} \times \mathcal{Y} \rightarrow \mathbb{R}\) the inter-space cost and \(\alpha \in [0,1]\) the trade-off parameter. For discrete settings, this can be solved by iteratively updating the coupling using conditional gradient updates~\citep{vayer2020contribution,sejourne2021unbalanced}.

\section{Primer on Flow matching}\label{appendix:flow_matching}
Given a smooth time-varying vector field \( v : [0, 1] \times \mathbb{R}^d \rightarrow \mathbb{R}^d \) we can define an ordinary differential equation (ODE):
\begin{equation}\label{equ:ode}
\frac{dx}{dt} = v_t(x)~,
\end{equation}
where the solution is a flow denoted by \( \phi_t(x)\) describing the trajectory of a point \( x \) over time with an initial condition \(\phi_0(x) = x\). The evolution of an initial probability distribution \( \rho_0 \in \mathcal{P}(\mathbb{R}^d)\) to a probability path \( p_t(x) \) under this flow is governed by the continuity equation:
\begin{equation}
 \frac{\partial p_t}{\partial t} = -\nabla \cdot (p_t v_t)
\end{equation}
The distribution \( p_t \) is the push forward of the initial distribution \( \rho_0 \) by the flow \( \phi_t \), denoted \( p_t = (\phi_t)_{\#} \rho_0 \), which describes how the distribution evolves under the influence of the flow. In Continuous Normalising Flows~\citep{chen2018neural} (CNFs), this vector field \( v_{t,\theta}(x) \) is parameterised using a neural network (\(\theta\)) that is optimised to satisfy the terminal condition \( \rho_1 = (\phi_t)_{\#} \rho_0 \), where \(\phi_t\) is a flow associated with the vector field.
\section{Related literature}\label{sec:related_literature}
\vspace{-0.5em}

Fusion techniques have been introduced to combine different modalities. For example, multi-modal encoders such as CLIP~\citep{radford2021clip} and AIGN~\citep{jia2021scaling} map data from distinct domains into a shared representation using a contrastive objective~\citep{oord2018contrastive}. These models often surpass traditional approaches in zero-shot transfer tasks on new datasets~\citep{chen2020big,kolesnikov2019large}, but typically require large amounts of paired data for training and come with substantial computational costs.

Model stitching~\citep{lenc2015understanding} represents another line of work, where intermediate latent representations from one model are transformed into another by learning a stitching module. This technique has been employed to align representations within models ~\citep{lenc2015understanding,csiszarik2021similarity} or across different modalities~\citep{merullo2022linearly}. While effective in learning transformations, this method generally requires end-to-end training of the stitching module, which is impractical when only the latent representations of the source and target models are accessible during training.

Within a semi-supervised learning setup, ~\cite{klebe2023gera} proposed to learn a shared embedding space by mapping the representations from two pre-trained multi-modal models into a common space. This approach necessitates only a small amount of labelled data but requires an additional model to be trained in the joint space for downstream tasks. In contrast, our approach directly identifies the transformation between the two latent representations, bypassing the need for a shared embedding space and additional models.



\section{Algorithm for learning multi-modal bridges}\label{appendix:architecture_algorithm}

We proposed model space alignment via multi-modal bridges using three different alignment strategies. For this, we align latent space distributions by either solving an OT problem (local and global) or true paired samples (true), and then learn flow matching for out-of-sample predictions (Algorithm~\ref{alg:m3}).
\vspace{1em}

\begin{algorithm}[H]
\caption{Learning inter-modal morphs via \textcolor{RedOrange}{true}, \textcolor{teal}{global}, or \textcolor{OliveGreen}{local} alignment}
\begin{algorithmic}
\State \textbf{Input:} 
\State ~~Two pre-trained models \(f_x\) and \(f_y\) 
\State ~~Two datasets \(D_\mathcal{X}\) and \(D_\mathcal{Y}\) \Comment{source and target domains}
\State ~~Paired samples \(P\) \Comment{optional paired samples}
\State ~~Entropy regularisation parameter \(\epsilon\) 
\State ~~\textcolor{blue}{Unbalanced weighting parameters  \(\tau = (\tau_{\mathcal{X}}, \tau_{\mathcal{Y}})\)}
\State ~~\textcolor{blue}{Reweighting
neural networks \(\eta_\theta\) and \(\xi_\theta\)} 
\State ~~Batch size \(b\)
\State ~~Number of iterations \(T_{\text{iter}}\)
\State ~~\(OT\) solver and cost function
\vspace{1mm}
\State \(X \gets f_x(D_\mathcal{X}), Y \gets f_y(D_\mathcal{Y})\) \Comment{Extract latent features using pre-trained models}
\vspace{1mm}
\State \textcolor{RedOrange}{\(\pi \gets \pi^{\text{true}}(X,Y,P)\)} \Comment{true alignment based on paired samples \(P\) using Eq.\ref{eq:true}}
\State \textcolor{teal}{\(C_{XY} \gets \text{fused\_cost}(X,Y,P)\)}
\State \textcolor{teal}{\(\pi \gets OT_{\epsilon, \textcolor{blue}{\tau}}(X,Y,C_{XY})\)} 
\vspace{1mm}
\For{\(t = 1, \dots, T_{\text{iter}}\)} 
    \State Sample \(x_1, \dots, x_b \sim X\) and \(y_1, \dots, y_b \sim Y\)
    \State \textcolor{OliveGreen}{\(C^b_{XY} \gets \text{fused\_cost}\left([x_i]_{i=1}^b, [y_i]_{i=1}^b,P\right)\)}
    \State \textcolor{OliveGreen}{\(\pi \gets OT_{\epsilon,\textcolor{blue}{\tau}}\left([x_i]_{i=1}^b, [y_i]_{i=1}^b,C^b_{XY}\right)\)} 
    \State Sample \((i_1, j_1), \dots, (i_b, j_b) \sim \pi\)
    \State Sample \(z_1, \dots, z_b \sim \mathcal{N}(0, 1)\), \(t_1, \dots, t_b \sim \mathcal{U}([0, 1])\) 
    \State \(\mathcal{L}(\theta) \gets \sum_{k} \left\|v_{t,\theta}([z_k,y_{j_k}]|t, x_{i_k}) - (y_{j_k} - z_k)\right\|_2^2\) +
    \State \quad \(\textcolor{blue}{\sum_k \left( \eta_{\theta}(\mathbf{x}_k) - b \pi_\mathcal{X}^k \right)^2 
+ \left( \xi_{\theta}(\mathbf{y}_k) - b \pi_\mathcal{Y}^k \right)^2}.\) 
    \State \(\theta \gets \text{Update}\left(\theta, \frac{1}{b} \nabla \mathcal{L}(\theta)\right)\) 
\EndFor
\end{algorithmic}
\label{alg:m3}
\end{algorithm}

\vspace{1em}

Algorithm~\ref{alg:m3} was implemented in \texttt{JAX}~\citep{jax2018github} using \texttt{Flax}~\citep{flax2020github}. For discrete OT solvers, we used the \texttt{OTT-JAX} library~\citep{cuturi2022ott}. To compute the KCCA between samples, we used the \texttt{MVLearn} library~\citep{perry2021mvlearn}. We used these default hyper-parameters for training -- unless explicitly specified otherwise:
\begin{itemize} 
 \item \textbf{optimiser}: adam (learning rate = \(10^{-4}\))
 \item \textbf{batch size} = \(256\)
 \item  \textbf{entropy regularisation \(\epsilon\)} = \(5\times10^{-3}\) with normalised cost matrices 
  \item  \textbf{fused penalty \(\alpha\)} = \(0.5\)
  \item \textbf{max number of iterations \(T_{iter}\)} = \(10,000\)
  \item \textcolor{blue}{\textbf{Unbalanced weighting parameters} \(\tau = (1,1)\)}
  \item \textcolor{blue}{\textbf{Reweighting
neural networks} \(\eta_\theta\), \(\xi_\theta\): multi-layer perceptron (MLP) used in \cite{klein2023generative}}.
\end{itemize}

\paragraph{Out-of-sample prediction} At inference time, we solved Eq.~\ref{equ:ode} for \( t_1 = 1 \) using the velocity field \( v_{t,\theta} \), a sampled point from the noise distribution \( z \) as the initial condition at \( t_0 = 0 \) conditioned on some out-of-sample point \( x \). 

The solution follows the form:
\begin{equation}\label{eq:pred}
    \hat{y} = \text{ODESolve}(v_{t,\theta}(\cdot \mid x), z, t_0=0, t_1=1),
\end{equation}
where the function \( \text{ODESolve} \) numerically solves the ODE from \( t_0 \) to \( t_1 \), yielding \( \hat{y} \) as the transported output at time \( t_1 = 1\), while the condition \(x\) modifies the evolution of the ODE as necessary.

\section{Learning the velocity field}\label{appendix:architecture} 

\begin{figure}[!t]
    \centering
    \includegraphics[width=0.45\linewidth]{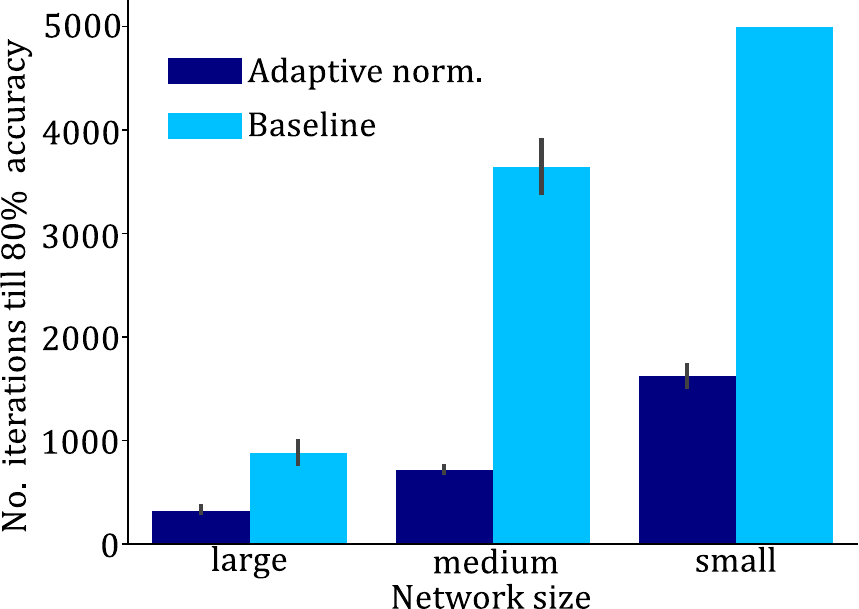}
    \caption{Iterations required to achieve \(80\%\) accuracy in MNIST experiments.
    \vspace{-1em}}
    \label{fig:iterations_arch}
\end{figure}
To learn the velocity field \(v_{t,\theta}\) two different architectures were considered: \(1)\) \textbf{MLP (i.e., Baseline):} three separate blocks for latent noise, time, and condition, which were concatenated and processed by a final MLP block following~\citep{klein2023generative}, and \(2)\) \textbf{adaLN:} blocks with adaptive layer normalisation (adaLN) following~\citep{perez2018film}. For each, we had three variations--small, medium, and large (Table~\ref{tab:vf_models})-- with the SiLU activation function applied after every layer in all models. For each setting, we measured the number of iterations required to achieve \(80\%\) accuracy in image-to-text experiments on the MNIST dataset, using true alignment in a supervised setting. The results demonstrate the effectiveness of the AdaLN architecture, which converges more rapidly and attains acceptable performance levels more efficiently (Figure \ref{fig:iterations_arch}). 
\begin{table}[!h]
    \centering  
    \begin{tabular}{llll}
        \toprule
        Model & Layers \(N\) & Hidden size \(d\) & Parameters \\
        \toprule
        MLP-Small & \(4\) & \(256\)  & \(700\)K  \\
        MLP-Medium & \(6\) & \(512\)  & \(3\)M \\
        MLP-Large & \(8\) & \(1680\) & \(49\)M  \\
        \midrule
        adaLN-Small & \(5\) & \(128\) & \(700\)K  \\
        adaLN-Medium & \(7\) & \(256\) & \(3\)M \\
        adaLN-Large & \(8\) & \(1024\) & \(49\)M  \\
        \bottomrule
    \end{tabular}%
    \caption{Different architectures for velocity field network\( v_{{t,\theta}} \). Here, MLP-X architecture refers to the \cite{klein2023generative} baseline. 
    \vspace{-1em}} 
    \label{tab:vf_models}
\end{table}
\subsection{Adaptive layer normalisation} \label{appendix:architecture_adap}
Diffusion Transformers (DiT)~\citep{peebles2023dit} and earlier works on diffusion models with U-net backbones~\citep{dhariwal2021diffusion} demonstrated the effectiveness of adaLN. For our formulation, we similarly replaced the standard MLP blocks with adaptive layer norm blocks. Unlike traditional layer normalisation, which directly learns the scale and shift parameters, adaLN regresses these parameters based on a time-dependent condition vector. The normalised output is then combined with the original input through a residual connection, where dimension-wise scaling parameters, initialised to zero, are applied. This adaptive mechanism enables more flexible and dynamic normalisation in response to the varying conditions during training.
\vspace{-0.5em}
\section{Fused costs}\label{appendix:fused_cost}

\paragraph{K-Nearest Neighbour cost} 

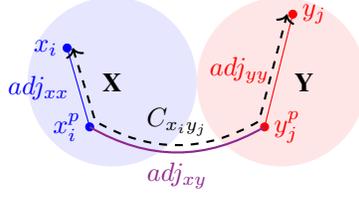
\begin{figure}[!t]
\centering
\begin{tikzpicture}[scale=0.75]
    \begin{scope}
        \fill[blue!10] (0,0) circle (1.5);
        \fill[red!10] (3,0) circle (1.5);
        \foreach \x/\y/\l in {-0.8/0.6/x_i, -0.4/-0.8/x^p_i}
        \filldraw[blue] (\x,\y) circle (2pt) node[left] {$\l$};
        \foreach \x/\y/\l in {3.2/1.2/y_j, 2.7/-0.8/y^p_j}
            \filldraw[red] (\x,\y) circle (2pt) node[right] {$\l$};
        \draw[blue, thin] (-0.8,0.6) -- (-0.4,-0.8) node[midway, left] {$adj_{xx}$};
        \draw[red, thin] (3.2,1.2) -- (2.7,-0.8) node[midway, left] {$adj_{yy}$};
        \draw[Plum, thick] (-0.4,-0.8) to[bend right] node[midway, below] {$adj_{xy}$} (2.7,-0.8);

        \draw[<->, thick, black, dashed] (-0.7,0.6) -- (-0.3,-0.7)  to[bend right] node[midway, above] {$C_{x_iy_j}$} (2.6, -0.7)  -- (3.1,1.2);
        
        \node at (0,0) {\textbf{X}};
        \node at (3.4,0) {\textbf{Y}};
    \end{scope}
    
\end{tikzpicture}
\caption[KNN cost]{Overview of KNN fused cost. We construct a fused graph from K-Nearest Neighbour (KNN) graphs for spaces $\mathcal{X}$ (blue) and $\mathcal{Y}$ (red). Intra-space adjacency matrices $\text{adj}_{XX}$ and $\text{adj}_{YY}$ and $\text{adj}_{XY}$ are used to form this graph. The fused graph allows estimation of the fused cost matrix $C^{knn}_{XY}$ via shortest path distances using a heat kernel approximation. }
\label{fig:knn}
\end{figure}

The use of K-Nearest Neighbour (KNN) graphs and the shortest path distances, induced by Euclidean distance, has been proposed as a means to approximate geodesic distances on data manifolds~\citep{crane2013geodesics}. Notably, several studies have demonstrated the effectiveness of this data-driven cost function~\citep{moon2018manifold,demetci2022scot,huguet2023geodesic}. Inspired by this approach, we first calculate the intra-space matrices $C_{XX}$ and $C_{YY}$ using Euclidean distance:
\begin{equation}
    C_{XX}[i,j] = |x_i - x_j|^2, \quad
    C_{YY}[i,j] = |y_i - y_j|^2,
\end{equation}
where $x_i,x_j \in \gX$ and $y_i,y_j \in \gY$ represent data points in their respective spaces. Based on this, we compute intra-domain K-Nearest Neighbour adjacency matrix:
\begin{equation}    
    adj_{XX}[i,j] = |x_i - x_j|^2. \quad
\end{equation}
we construct an inter-space graph to approximate the fused-cost function. Given two intra-space k-nearest neighbour (kNN) adjacency matrix, $adj_{XX}$ and $adj_{YY}$, and a paired set $P$, we define the inter-space graph $G_{\text{fused}}$ as:
\begin{equation}
    G_{\text{fused}} = \text{graph\_from\_adj}
    \begin{bmatrix}
        \text{adj}_{XX} & \text{adj}_{XY} \\
        \text{adj}_{XY} & \text{adj}_{YY}
    \end{bmatrix}
\end{equation}
where $\text{adj}_{XX}$ and $\text{adj}_{YY}$ are the adjacency matrices of $G_{XX}$ and $G_{YY}$, respectively. The matrix $\text{adj}_{XY}$ is:
\begin{equation*}
    \text{adj}_{XY}[i,j] = 
\begin{cases}
1 & \text{if } (x_i,y_j) \in P \\
0 & \text{otherwise}
\end{cases}
\end{equation*}
Then, the fused cost matrix $C^{knn}_{XY}$ is the shortest path in $G_{\text{fused}}$: 
\begin{equation}
    C^{knn}_{{XY}}[i,j] = \text{ShortestPath}(G_{\text{fused}}, x_i, y_j)~,
\end{equation}
estimated using the heat kernel in our experiments~\citep{crane2013geodesics, heitz2021ground}. 

The heat kernel provides an approximation of the shortest path by modelling heat diffusion across the graph. Nodes that are closer in terms of the shortest path will exhibit faster heat diffusion, which allows us to estimate distances between them based on the behaviour of the heat kernel for small diffusion times.

\paragraph{Kernel canonical correlation analysis cost}
\begin{figure}[!t]
\centering
\begin{tikzpicture}[scale=0.75]
    \begin{scope}
        \fill[blue!10] (0,0) circle (1.5);
        \fill[red!10] (3,0) circle (1.5);
        \fill[cyan!5] (1.5,2) ellipse (2 and 1);
        \foreach \x/\y/\l in {-0.5/0.5/x_i}
            \filldraw[blue] (\x,\y) circle (2pt) node[below] {$\l$};
        \foreach \x/\y/\l in {3.5/-0.5/y_j}
            \filldraw[red] (\x,\y) circle (2pt) node[below] {$\l$};
        \foreach \x/\y/\l in {0.7/1.7/u_{xi}}
            \filldraw[blue] (\x,\y) circle (2pt) node[above] {$\l$};
        \foreach \x/\y/\l in {3/2.3/u_{yj}}
            \filldraw[red] (\x,\y) circle (2pt) node[above] {$\l$};
        \draw [Plum, thick, ->] (-0.5,0.5) to[bend left]  node[midway, left] {$\alpha$} (0.7,1.7);
        \draw [Plum, thick, ->] (3.5,-0.5) to[bend right] node[midway, right] {$\beta$} (3,2.3);
        
        \draw [cyan, thick, <->] (1,2.44) -- (2,2.44) node[midway, above, font=\tiny] {max. corr.};
        
        \draw[->, thick, black, dashed] (0.7,1.7) -- node[midway, below] {$C_{x_iy_j}$} (3,2.3);
        \node at (0,0) {\textbf{X}};
        \node at (3,0) {\textbf{Y}};
    \end{scope}

\end{tikzpicture}
\caption[KCCA cost]{Overview of KCCA fused cost. Using paired samples, we find the projection vectors $\alpha$ and $\beta$ to a joint space. The points $x_i$ and $y_j$ are then transformed into this joint correlation-maximising space, denoted as $u_{xi}$ and $u_{yj}$. Finally, we compute the distance between the points in this joint space.}
\label{fig:kcca}
\end{figure}
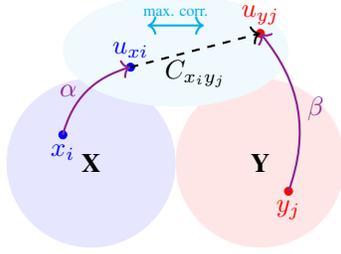

Kernel canonical correlation analysis~\citep{hardoon2004canonical} (KCCA) extends canonical correlation analysis (CCA)~\citep{hotelling1992relations} by projecting the data into a high-dimensional feature space using some kernel function $ k(\cdot,\cdot) $. It then finds projections in this high-dimensional space that maximise the correlation between the two sets of data. Using the paired point matrices $ X_p = [ x_{ip} ]_{p=1}^{l} $ and $ Y_p = [ y_{jp} ]_{p=1}^{l} $, where $ (x_{ip},y_{jp}) \in P $, we calculate the projection vectors $ \alpha $ and $ \beta $~\citep{hardoon2004canonical}:
\begin{equation}
    \rho = \max_{\alpha, \beta} \frac{\alpha K_{X_p} K_{Y_p} \beta}{\sqrt{\alpha K_{X_p}^2 \alpha \cdot \beta K_{X_p}^2 \beta}}.
\end{equation}
where these kernel matrices $ K_{X_p}= k(X_p,X_p) $ and $ K_{Y_p} = k(Y_p,Y_p) $ based on paired samples and a kernel function $k(\cdot,\cdot)$. we define the fused cost $ C_{XY}^{kcca} $ as:
\begin{equation}
\begin{aligned}
    u_{X} &= K_{X} \alpha \\
    u_{Y} &= K_{Y} \beta \\
    C_{XY}^{kcca}[i,j] &= \text{cosine\_distance}(u_{X}^{(i)}, u_{Y}^{(j)})~,
\end{aligned}
\end{equation}
where $u_{X}$ and $u_{Y} $ are the projections of the joint embedding space for the entire dataset. We used paired samples to compute the projection vectors $\alpha$ and $\beta$. 
For our experiments, KCCA is formalised using a Gaussian RBF kernel and paired points. By maximising the correlation between the projected variables from these two sets, KCCA tries to find a joint space for the maximum possible shared information.

\section{Models}\label{appendix:models}

\subsection{MNIST Experiments}

Here, we provide details about the different models considered for text-image alignment using the MNIST dataset; including the pre-trained models (\textit{VAE\(_\text{image}\)} and \textit{VAE\(_\text{text}\)}) and the fusion baseline (Fused VAE)~\citep{Kingma2014vae}:

\subsubsection{\textit{VAE\(_\text{image}\)} and \textit{VAE\(_\text{text}\)}}\label{appendix:vae-pre-trained} \textit{VAE\(_\text{image}\)} was trained on \(50,000\) image samples of size (\(28,28,1\)) from the MNIST training dataset, using mean binary cross-entropy as the reconstruction loss. The image pixel values were converted to \(1\) if the value was higher than \(0.5\), otherwise to \(0\). Separately, \textit{VAE\(_\text{text}\)} was trained to compress one-hot encoded labels into a compact space and reconstruct the original labels from the input, and was trained using softmax cross entropy as the reconstruction loss. Table~\ref{tab:arch-pre-trained} provides architecture details. 

\begin{table}[!h]
\centering
\begin{tabular}{cl}
\toprule
\textbf{Model Type}    & \textbf{Architecture}              \\ 
\toprule
\textit{VAE\(_\text{text}\)}  & 
    \begin{tabular}[c]{@{}l@{}}
        Input: \((10)\) \\ 
        Encoder: FC \(64,32\) \\ 
        Latents: \(4\) \\ 
        Decoder: FC \(32,16,10\), softmax output
    \end{tabular} \\ 
\hline
\textit{VAE\(_\text{image}\)} & 
    \begin{tabular}[c]{@{}l@{}}
        Input: \((28\times28\times1)\) \\ 
        Encoder: Conv \(128,256,512\) \\ 
        Latents: \(16\) \\ 
        Decoder: ConvT \(256, 128, 1\), sigmoid output
    \end{tabular} \\ 
\bottomrule \\
\end{tabular}
\caption{\textit{VAE\(_\text{image}\)} and \textit{VAE\(_\text{text}\)} architectures. After each layer, we apply a ReLU non-linearity. For the convolutional (conv) layers, we used \(3 \times 3\) kernels, with strides set to (2, 2) and \textit{same} padding consistently across all models.
}
\label{tab:arch-pre-trained}
\end{table}

\vspace{-0.5em}
\subsubsection{VAE baselines} 
The baseline models -- depending on the task were trained in an end-to-end fashion -- for either reconstructing images from text using binary cross-entropy or reconstructing labels from image input using the softmax cross-entropy function. Each of these baselines was trained using different latent dimensions (large \(=128\) and small \(=16\); Table~\ref{tab:arch-baselines}) 

For the Fused VAE, we modified the vanilla VAE model to have two separate encoders and decoders for each modality. For this model, the encoders learn to map data from different modalities to a joint embedding space, and each decoder reconstructs the output based on the representation in this joint space. To construct the Fused VAE, we modified the ELBO:
\begin{equation}
\begin{aligned}
    \mathcal{L}_{\text{ELBO}-{q_{\phi1}}}= \underbrace{\mathbb{E}_{q_{\phi1}(\mathbf{z}|\mathbf{x})}\left[\log p_{\theta1}(\mathbf{x}|\mathbf{z})\right]}_{\text{Reconstruction Loss 1}} + \underbrace{\mathbb{E}_{q_{\phi1}(\mathbf{z}|\mathbf{x})}\left[\log p_{\theta2}(\mathbf{y}|\mathbf{z})\right]}_{\text{Reconstruction Loss 2}}
    - \underbrace{\text{KL}(q_{\phi1}(\mathbf{z}|\mathbf{x}) \parallel p(\mathbf{z}))}_{\text{KL Divergence }},
    \end{aligned}
\end{equation}
where \( q_{\phi_1} \) and \( q_{\phi_2} \) represent the distributions parameterisations by \( \phi_1 \) and \( \phi_2 \) for the first and second encoders, respectively. Similarly \( p_{\theta_1} \) and \( p_{\theta_2} \) denote the distributions parameterised by \( \theta_1 \) and \( \theta_2 \) for the first and second decoders. Given these, Fused VAE loss was defined as:
\begin{equation}
    \mathcal{L}_{\text{ELBO-fused}}= \mathcal{L}_{\text{ELBO}-{q_{\phi1}}} + \mathcal{L}_{\text{ELBO}-{q_{\phi2}}}~.
\end{equation}
We trained two variants of the Fused VAE using paired samples from the two domains; Fused VAE (E2E) was trained on the raw data Fused VAE (Latent) was trained on the latent spaces of pre-trained (\textit{VAE\(_\text{image}\)}) and (\textit{VAE\(_\text{text}\)}). This allows for a nuanced comparison of how each model variant handles multi-modal data integration.
\begin{table}[!h]
\centering
\begin{tabular}{ccl}\toprule
\textbf{Model Type} & \textbf{Task} &\textbf{Architecture} \\
\toprule
 \textit{VAE (small)} & text->image & \begin{tabular}[c]{@{}l@{}}
            Input: \((10)\)\\
            Encoder: FC \(64,32\).\\ 
            Latents: \(16\)\\
            Decoder: ConvT \(1024,512,1\), sigmoid output.
        \end{tabular} \\ \hline
 \textit{VAE (large)} & text->image & \begin{tabular}[c]{@{}l@{}}
            Input: \((10)\)\\
            Encoder: FC \(64,32\).\\ 
            Latents: 128\\
            Decoder: ConvT 1024,512,1, sigmoid output.
        \end{tabular} \\ \midrule

\textit{VAE (small)} & image->text  & \begin{tabular}[c]{@{}l@{}}
            Input: \((28\times28\times1)\)\\
            Encoder: Conv \(512, 1024, 2048\).\\ 
            Latents: 16\\
            Decoder: FC \(32,16, 10\), softmax output.
        \end{tabular}\\ \hline

\textit{VAE (large)} & image->text & \begin{tabular}[c]{@{}l@{}}
            Input: \((28\times28\times1)\)\\
            Encoder: Conv \(512, 1024, 2048\).\\ 
            Latents: \(128\)\\
            Decoder: FC \(32,16, 10\), softmax output.
        \end{tabular}\\ \hline
        
\textit{Fused VAE (E2E)} & image<->text  & \begin{tabular}[c]{@{}l@{}}
            Input1: \((28\times28\times1)\)\\
            Input2: \((10)\)\\
            Encoder1: Conv \(512, 1024, 2048\).\\ 
            Encoder2: FC \(64,32\).\\ 
            Latents: \(128\)\\
            Decoder1: ConvT \(1024,512,1\), sigmoid output. \\
            Decoder2: FC \(32,16, 10\), softmax output.
        \end{tabular}\\ \hline

\textit{Fused VAE (latent)} & image<->text & \begin{tabular}[c]{@{}l@{}}
            Input1: \((16)\)\\
            Input2: \((4)\)\\
            Encoder1: FC \(8\times1024\) \\ 
            Encoder2: FC \(8\times1024\)\\ 
            Latents: 128\\
            Decoder1: FC \(8\times1024,16\). \\
            Decoder2: FC \(8\times1024,4\).
        \end{tabular}\\

\bottomrule
\\
\end{tabular}
\caption{Baseline models architectures for the MNIST experiments. Here, FC is for the fully connected layer, Conv is for the convolutional layer and ConvT is for the convolutional transpose layer.}
\label{tab:arch-baselines}
\end{table}

\subsection{ImageNet Experiments}
\subsubsection{Vision Transformer}
The Vision Transformer (ViT)~\citep{dosovitskiy2021imageworth16x16words} leverages the transformer~\citep{vaswani2017attention} architecture to process images by dividing them into patches and applying self-attention to capture global relationships. The original paper introduced three variants of ViT: ViT-B (Base), ViT-L (Large), and ViT-H (Huge), each differing in scale and complexity. For feature extraction for ImageNet, we used ViT-B model pre-trained~\citep{wu2020visual} on ImageNet-21k~\citep{deng2009imagenet} (which contains \(14\) million images and \(21,843\) classes) at a resolution of \(224 \times 224\). For image pre-processing, we followed the procedure outlined in~\cite{dosovitskiy2021imageworth16x16words}.

\subsubsection{Sentence Transformer}\label{appendix:sent-trans}
The Sentence Transformer~\citep{reimers-2019-sentence-bert}, commonly known as SBERT, converts sentences and paragraphs into embeddings that capture the high-level semantic meaning of the text. A common application of sentence transformers is measuring semantic similarity between sentences using cosine similarity. For our experiments, we used the pre-trained model \texttt{all-MiniLM-L6-v2} from the Hugging Face repository, which is based on the MiniLM architecture~\citep{wang2020minilm}, to extract textual features from the input prompts.

\subsection{\cite{majaj2015simple} Experiments}\label{appendix:majaj}
\subsubsection{Artificial neural networks}

For our experiments, we selected six pre-trained neural networks. Among them, we used three publicly available convolutional neural network (CNN) architectures, VGG16~\citep{simonyan2014very}, ResNet50~\citep{he2016deep}, and EfficientNet-B0~\citep{tan2019efficientnet}, all pre-trained on the ImageNet dataset, and the weights were obtained from the PyTorch library~\citep{paszke2019pytorch}. These models were selected for their high Brain-Score performance (\url{https://www.brain-score.org/benchmark/vision/44}). In addition, we included the brain-inspired CORNet-S architecture (weights from \url{https://github.com/dicarlolab/CORnet}), the adversarially robust Robust-ResNet50-L2-3 model (\(\ell_2\) norm, \(\epsilon=3\); weights from \url{https://github.com/MadryLab/robustness}), and a Vision Transformer ViT-L-32~\citep{dosovitskiy2021imageworth16x16words}.

For all CNNs, we extracted activations from each ReLU nonlinearity following every convolutional layer for each stimulus, as in~\citep{canatar2024spectral}. For ViT, we collected activations from all intermediate encoder layers. Following \citep{SchrimpfKubilius2018BrainScore}, we kept the first $1000$ principal components per layer using $1000$ validation images. For all further analysis, we selected best three layers based on their performance in neural predictivity (Brain-Score)\citep{SchrimpfKubilius2018BrainScore}, effective dimensionality (ED)\citep{elmoznino2024high}, and representational similarity analysis (RSA)~\citep{kriegeskorte2008representational} for IT and V4 regions. Table~\ref{tab:arch-majaj} summarizes each model’s ImageNet Top-1 accuracy and the layers chosen for feature extraction.

\begin{table}[!h]
  \centering
  \begin{tabular}{l c l}
    \toprule
    \textbf{Model} & \textbf{ImageNet Top-1} & \textbf{Selected layers} \\
    \midrule
    VGG16 & $71.3\%$ & 
      \begin{tabular}[t]{@{}l@{}}
        \textit{features.13}, \textit{features.15}, \textit{features.22}\\
        \textit{features.25}, \textit{features.27}, \textit{features.29}
      \end{tabular} \\
    \midrule
    ResNet50 & $80.63\%$ &
      \begin{tabular}[t]{@{}l@{}}
        \textit{layer2.2}, \textit{layer2.3}, \textit{layer3.4}, \textit{layer3.5}\\
        \textit{layer4.0}, \textit{layer4.1}, \textit{layer4.2}
      \end{tabular} \\
    \midrule
    EfficientNet-B0 & $76.87\%$ &
      \begin{tabular}[t]{@{}l@{}}
        \textit{layers.1}, \textit{layers.2}, \textit{layers.3}\\
        \textit{layers.5}, \textit{layers.6}, \textit{features}
      \end{tabular} \\
    \midrule
    ViT-L-32 & $76.35\%$ &
      \begin{tabular}[t]{@{}l@{}}
        \textit{Encoder.layer\_10}, \textit{Encoder.layer\_11} \\
        \textit{Encoder.layer\_12}, \textit{Encoder.layer\_17} \\
        \textit{Encoder.layer\_18}, \textit{Encoder.layer\_19}
      \end{tabular} \\
    \midrule
    CORNet-S & $72.51\%$ &
      \begin{tabular}[t]{@{}l@{}}
        \textit{module.V1}, \textit{module.V2}, \textit{module.V4}, \textit{module.IT}
      \end{tabular} \\
    \midrule
    Robust-ResNet50-l2-3 & $72.51\%$ &
      \begin{tabular}[t]{@{}l@{}}
        \textit{layer2.0}, \textit{layer2.1}, \textit{layer2.2}, \textit{layer2.3}\\
        \textit{layer3.4}, \textit{layer3.5}, \textit{layer4.0}, \textit{layer4.1}\\
        \textit{layer4.2}
      \end{tabular} \\
    \bottomrule
  \end{tabular}
  \caption{List of selected models with their ImageNet Top-1 accuracies and the corresponding layers used for feature extraction.}
  \label{tab:arch-majaj}
\end{table}

\section{Metrics}\label{appendix:metrics}
Here, we introduce the different metrics used in Sec.~\ref{sec:results}.
\vspace{-0.5em}

\subsection{Feature overlap}\label{appendix:overlap}
We compute the overlap in the feature space, assuming the feature space is Euclidean, for a pre-trained model using a metric derived from k-nearest neighbours (kNN). We randomly select a batch of $b$ samples, $ X = \{x_i\}_{i=1}^b $, from the domain $ \gX $, with corresponding labels $ L = \{l_i\}_{i=1}^b $. First, we calculate the $k$-nearest neighbours for each point $ x_i $ in the feature space. For each point $ x_i $, the overlap is defined as the proportion of its $k$-nearest neighbours that have a label different from $ l_i $. If $ \psi_{knn}(x_i) $ represents the set of the $k$-nearest neighbours of $ x_i $, then the overlap for a given point $ x_i $ is:

\begin{equation}
\text{overlap}(x_i) := \frac{1}{k} \sum_{x_j \in \psi_{knn}(x_i)} \mathds{1}(l_j \neq l_i)~,
\end{equation}

where $\mathds{1}(l_j \neq l_i) $ is the indicator function, which equals 1 if the label $ l_j $ of the neighbour $ x_j $ differs from the label $ l_i $, and 0 otherwise. We approximate the feature space overlap as the mean of the individual overlaps for all samples $ x_i $ in the batch. This metric provides a measure of how often points in the feature space are surrounded by neighbours with different labels, reflecting the degree of separation within the feature space. For all experiments reported in Sec.~\ref{sec:results}, we used $k=15$ and five randomly selected batches, each with size $b=100K$ for ImageNet experiment and the $b=5K$ for MNIST expriments.

\subsection{Evaluating MNIST}\label{appendix:mnist}
\textbf{Image-to-text } We use the reconstruction cost as the reported test accuracy for the VAE and fused-VAE (E2E) baseline models. For models that rely on latent spaces, such as morphing models and latent VAE, after transforming the test split images from the \textit{VAE$_\text{image}$} to the \textit{VAE$_\text{text}$} latent space, we reconstruct the labels using the \textit{VAE$_\text{text}$} decoder and report this as the test accuracy.

\textbf{Text-to-image:} For the baseline models with end-to-end training we construct images from the labels. However, for other models that transfer representations between latent spaces, we use the respective transformed representations and construct an image using \textit{VAE$_\text{image}$}. Assuming we have $n$ text-image pairs $\{(x_i, y_i)\}_{i=1}^n$ in the test dataset, where each $x_i \in \{0,1\}^{28 \times 28}$, and let $\{\Tilde{x_i}\}_{i=1}^n$ be the samples reconstructed using the transportation method $T$. Assume that $S_c$ is the set of indices such that $i \in S_c \Rightarrow y_i = c$ and $|S_c| = N_c$. We construct the $2$D pixel-wise distributions for the original and reconstructed images for each class $c$: 
\begin{equation}
\begin{aligned}
P(X^{j,k} \mid c) = \frac{1}{N_c} \sum_{i \in S_c} x_i^{j,k}, \\
P(\Tilde{X}^{j,k} \mid c) = \frac{1}{N_c} \sum_{i \in S_c} \Tilde{x}_i^{j,k},
\end{aligned}
\end{equation}
where $X^{j,k}$ represents the pixel at position $(j,k)$ in the original images, and $\Tilde{X}^{j,k}$ represents the corresponding pixel in the reconstructed images. 
From this, we calculate the  Next, we define the mean squared error (MSE) metric between the pixel-wise distributions for the original images ($X$) and the reconstructed ones ($\Tilde{X}$) for class $c$, denoted as $MSE(c)$:
\begin{equation}
    MSE(c) = \frac{1}{28^2} \sum_{j=1}^{28}\sum_{k=1}^{28} \left( P(X^{j,k} \mid c) - P(\Tilde{X}^{j,k} \mid c) \right)^2.
\end{equation}
Finally, the mean squared error for the transportation function is defined as:

\begin{equation}
    MSE = \frac{1}{10} \sum_{c=0}^{9} MSE(c).
\end{equation}

\subsection{Evaluating ImageNet}\label{appendix:imagenet-acc-metric}

For the ImageNet experiments, after training the velocity field $ v_{t,\theta} $, for each image $ d_j^x $ with label $ l_j $ in the test split, we compute its representation $ x_j $ in the embedding space of the pre-trained image model. We then use Eq.\ref{eq:pred} to obtain the corresponding prediction $ \hat{y}_j $ in the target space. 

The ImageNet dataset contains 1000 unique classes, and we assume their representations in the embedding space of the language domain are denoted as $ Y = [y_i]_{i=1}^{1000} $, corresponding to the labels $ [l_i]_{i=1}^{1000} $. To classify each predicted point $ \hat{y}_j $, we compute the cosine distance between $ \hat{y}_j $ and all points $ y_i $ in the target space. The nearest neighbour $ y_{l_k} $ is the point that minimises cosine distance:
\begin{equation}
    k = \arg\min_{i} \text{cosine\_distance}( \hat{y}_j, y_i )
\end{equation}

where $ l_k $ is the label corresponding to the closest. The accuracy is then computed as follows:

\begin{equation}
    \text{accuracy} = \frac{1}{n} \sum_{j=1}^{n} \mathds{1}(l_j = l_k)
\end{equation}

where $\mathds{1}$ is the indicator function and $n$ is the total number of test samples.

\subsection{Calculating intra-space cost for \cite{majaj2015simple}}\label{appendix:inter-space_majaj}
To compute the intra-space cost matrix $C_{XX}$ for neural activity responses, we use a correlation-driven cost similar to  ~\cite{yamins2014performance}. Let $x_i, x_j \in \mathcal{X}$ be the neural responses to stimuli $s_i$ and $s_j$, respectively. We define the cost matrix as:

\begin{equation}
    C_{XX}(x_i, x_j) = 1 - \frac{\text{cov}(x_i, x_j)}{\sqrt{\text{var}(x_i) \text{var}(x_j)}}.
\end{equation}

Similar to previous experiments, we normalise the intra-space cost matrix by its mean before computing the optimal coupling.

\subsection{Artificial-to-biological neural representation evaluation}\label{appendix:majaj_svm}

Following \cite{majaj2015simple, kar2019evidence}, we evaluated the morphs' performance in a category object classification task (Fig.~\ref{fig:majaj_class}) using support vector machine (SVM) classifiers~\cite{chang2011libsvm}. We did not apply neural re-weighting (Eq.~\ref{eq:genot}) to compute the mappings. For V4 and IT region, we trained separate SVMs to decode neural activity corresponding to the category of core images using the entire training dataset. We employed a C-Support Vector Classification(C-SVC) model with a linear kernel and hinge loss with \(L_2\) regularisation, and performed $5$-fold cross-validation for hyperparameter optimisation.
\section{GENOT validation}\label{appendix:val_example}
\begin{figure}[!t]
    \centering
    \includegraphics[width=0.8\linewidth]{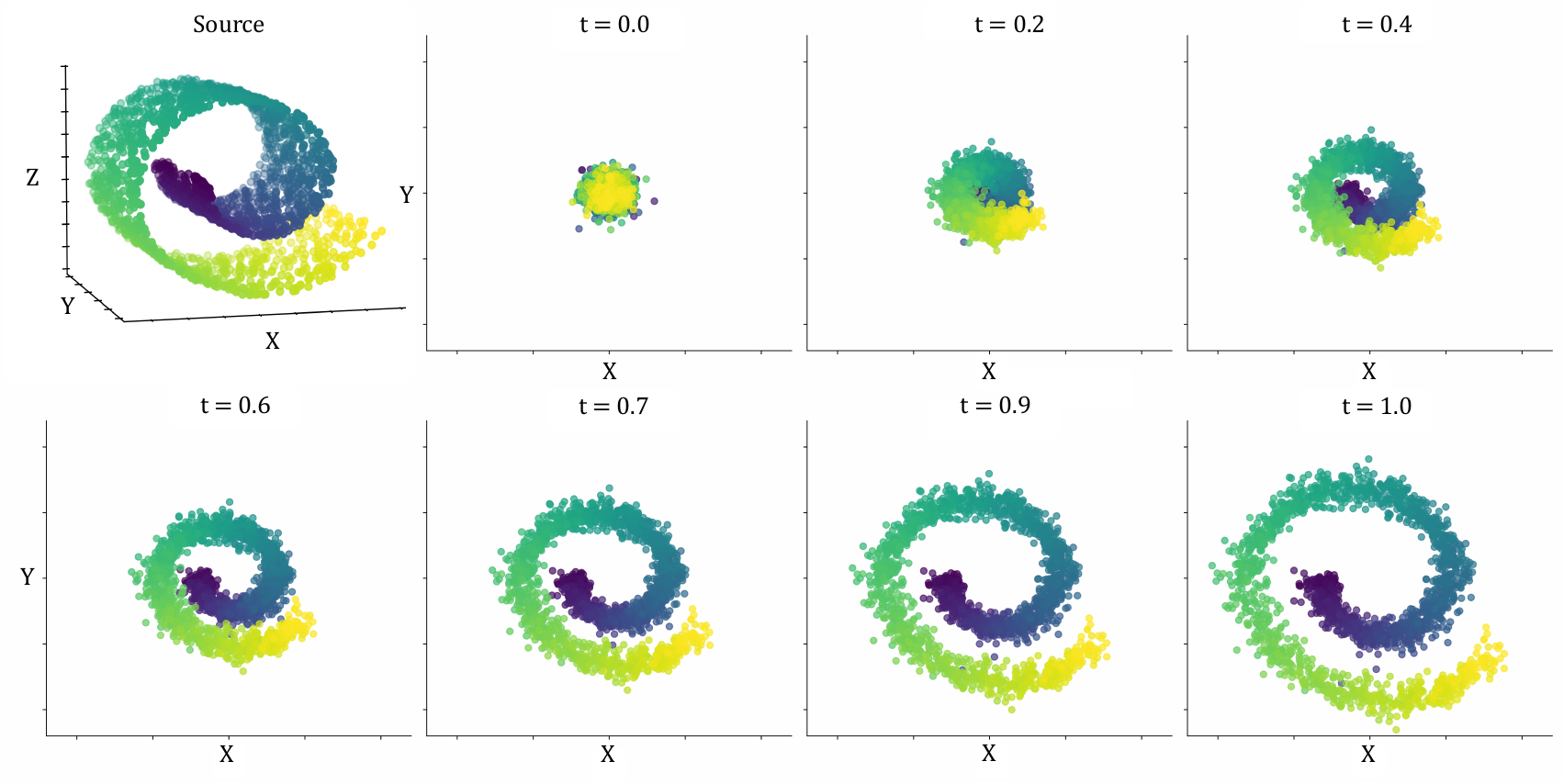}
    \caption{Mapping from Swiss roll to Spiral using local alignment}
    \label{fig:swirl}
\end{figure}
To ensure that baselines were consistent with reported results in~\cite{klein2023generative}, we replicate the results for the Swiss roll (\(\mathbb{R}^3\); source distribution) to spiral (\(\mathbb{R}^2\); target distribution) using the GW solver (unsupervised) and local alignment. Fig.\ref{fig:swirl} shows the evolution of the noise distribution into the target distribution in \(\mathbb{R}^2\) space.

\section{Time complexity of alignment strategies}\label{appendix:time_complexity}
\begin{figure}[!t]
    \centering
    \includegraphics[width=0.5\linewidth]{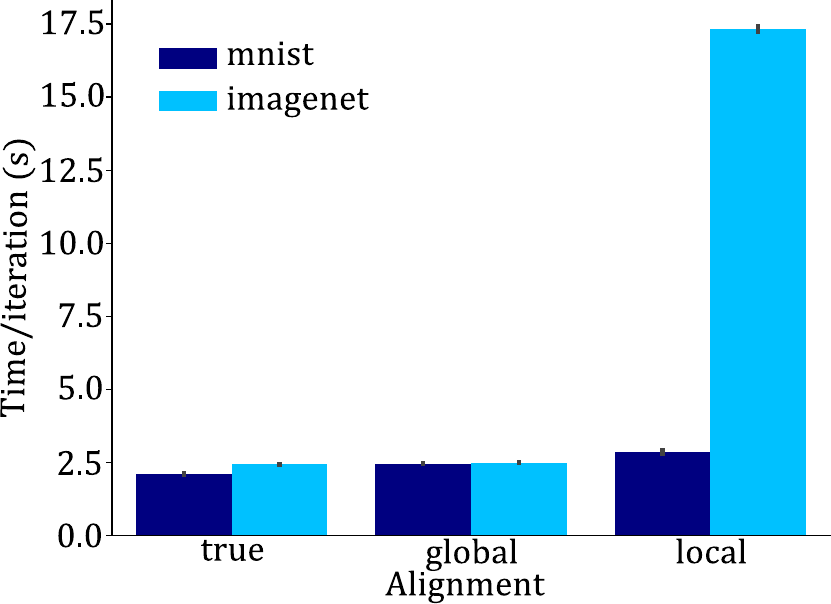}
    \caption{Runtime per iteration (in seconds) for different alignment strategies across MNIST and ImageNet datasets.}
    \label{fig:time_complexity}
\end{figure}
\vspace{-1em}
We evaluated the runtime per iteration for various alignment strategies (Fig.\ref{fig:time_complexity}) using the same computational setup. We observed that local alignment exhibits significantly higher time complexity compared to other strategies. This is primarily due to the necessity of solving an OT problem at each iteration (Algorithm~\ref{alg:m3}). In scenarios involving more complex distributions, such as those in the ImageNet dataset, the difference in computational cost becomes even more pronounced. The need to compute the fused cost and solve the OT problem in each iteration further exacerbates the time complexity in such cases.

\section{OT solver optimisation }\label{appendix:alpha_opt}
We examined the effectiveness of different OT discrete solvers in learning the optimal coupling, using large sample sizes ($100,000$ for ImageNet and $5,000$ for MNIST). Both global and local alignment strategies rely on OT discrete solvers, which can be either linear or fused FGW, and the performance of these solvers has a significant impact on the morphing quality. The FGW solver integrates intra-domain costs, \(C_{XX}\) and \(C_{YY}\), derived using cosine distance, and an inter-domain cost, which is computed based on the formulation introduced in Sec.\ref{sec:method-m3-bridge} and Appendix~\ref{appendix:fused_cost}. Additionally, the hyperparameter $\alpha$ controls the trade-off between quadratic and linear OT objectives.

For the ImageNet and MNIST experiments, we evaluated the matching accuracy of the discrete solvers as a function of the ratio of paired samples, varying both the values of $\alpha$ and the inter-domain cost function. In accordance with Eq.~\ref{eq:fgw}, as $\alpha \rightarrow 1$, the quadratic costs dominate over the linear component, favouring a solution that leans towards unsupervised alignment, as paired samples are used exclusively in the inter-domain cost. Conversely, when $\alpha = 0$, the problem simplifies to a linear OT problem.

\begin{figure}[!t]
    \centering
    \includegraphics[width=0.8\linewidth]{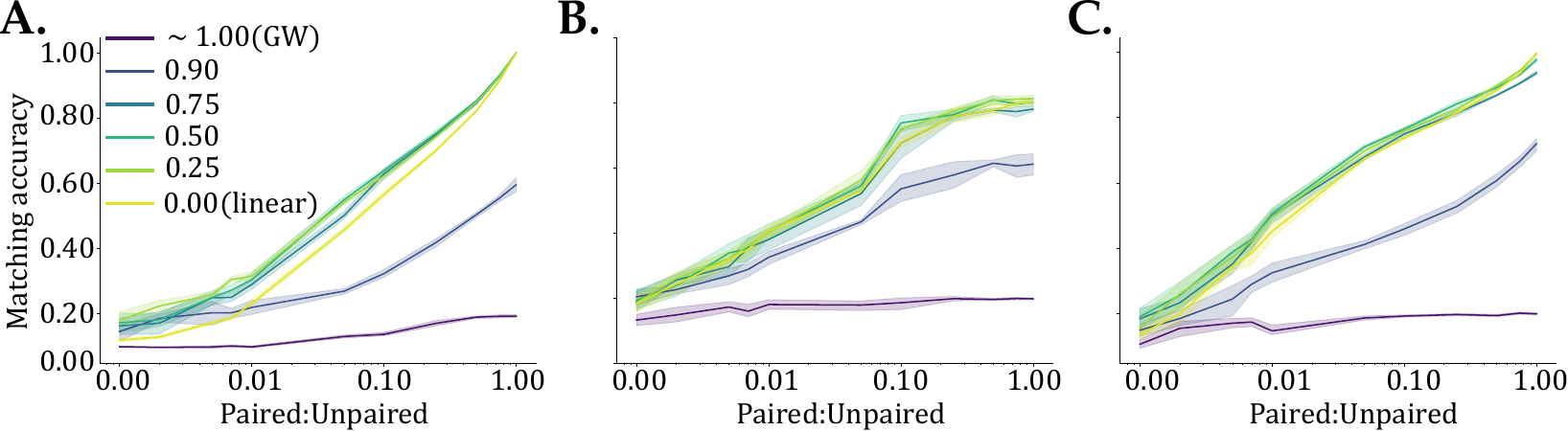}
    \caption[FGW solver optimisation for MNIST experiment]{Matching accuracy across different values of $\alpha$ for discrete OT solvers in the MNIST experiment. Accuracy was computed by evaluating the correct matches from the optimal coupling $\pi^\star$ for different fused costs. \textbf{A)} KNN, \textbf{B)} KCCA, \textbf{C)} Bridge.}
    \label{fig:alpha_mnist}
\end{figure}   
\begin{figure}[!h]
    \centering
    \includegraphics[width=0.8\linewidth]{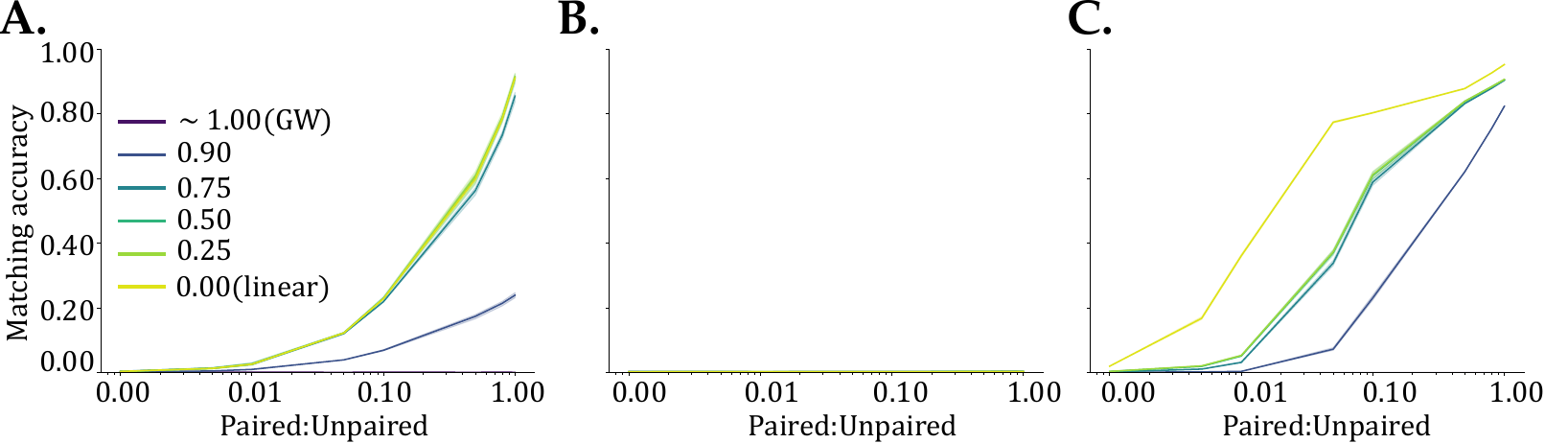}
    \caption[FGW solver optimisation for ImageNet experiment]{Matching accuracy across different values of $\alpha$ for discrete OT solvers in the ImageNet experiment. Accuracy was computed by evaluating the correct matches from the optimal coupling $\pi^\star$ for different fused costs. \textbf{A)} KNN, \textbf{B)} KCCA, \textbf{C)} Bridge.}
    \label{fig:alpha_imagenet}
\end{figure}

\section{Unbalanced Setting}\label{appendix:unbalanced}

\subsection{Theoretical Background For \textcolor{blue}{U}-Genot }\label{appendix:unbalanced_theory}

\textcolor{blue}{Unbalanced} optimal transport (\textcolor{blue}{U}-OT) is an extension of the classical OT problem, where the marginals of the optimal coupling \(\pi^\star\) found in Eqs.~\ref{eq:LEOT}-\ref{eq:fgw} can differ from the true source (\(\mu\)) and target (\(\nu\)) distributions~\citep{sejourne2023unbalanced}. By relaxing the constraint that mass must be exactly preserved between distributions, \textcolor{blue}{U}-OT can ignore or down-weight outliers and noisy samples that would otherwise force suboptimal transport plans. This makes \textcolor{blue}{U}-OT particularly well-suited for aligning neural activity data, which often contains measurement noise.

The unbalanced weighting parameters, \textcolor{blue}{\(\lambda_\mathcal{X}\)} and \textcolor{blue}{\(\lambda_\mathcal{Y}\)}, control the extent to which the marginals of the optimal coupling can diverge from the true source and target distributions. We follow the convention used by \cite{klein2023generative} to define: 

\begin{equation}\textcolor{blue}{
\tau_\mathcal{X} = \frac{\lambda_\mathcal{X}}{\lambda_\mathcal{X} + \epsilon}, \quad  
\tau_\mathcal{Y} = \frac{\lambda_\mathcal{Y}}{\lambda_\mathcal{Y} + \epsilon}}~,
\end{equation}

where we recover the classical OT problem by setting \(\textcolor{blue}{\tau_{i}} = 1\) when \(\textcolor{blue}{\lambda_{{i}}} \to \infty\). We note that the unbalancedness parameter \(\tau_i\) is influenced by the entropy regularisation parameter \(\epsilon\) in this definition. 

When using \textcolor{blue}{unbalanced} OT solvers in \textcolor{blue}{U}-GENOT, two reweighting functions, \textcolor{blue}{$\eta: \mathcal{X} \rightarrow \mathbb{R}^+$} and \textcolor{blue}{$\xi:  \mathcal{Y} \rightarrow \mathbb{R}^+$}, are employed for the source and target space, respectively. These reweighting functions are defined as \textcolor{blue}{$\pi^\star_{\mathcal{X}} = \eta \cdot \mu$} and \textcolor{blue}{$\pi^\star_{\mathcal{Y}} = \xi \cdot \nu$} in the unbalanced setting. Practically, these functions can be approximated by parameterising neural reweighting functions, \textcolor{blue}{$\eta_\theta$} and \textcolor{blue}{$\xi_\theta$}, which are trained to re-balance the \textcolor{blue}{U}-OT using Eq.~\ref{eq:genot}.

\subsection{Experimental considerations for aligning neural representations}\label{appendix:unbalanced_exp}
For the neural activity model experiments presented in Sec.\ref{sec:results::bio-art}, we used the unbalanced OT setting to account for noise in the neural recordings. However, the choice of unbalanced weighting parameters was empirically determined. In our case, since we are using a global strategy (with a low number of data points and low memory requirements), we solve the OT problem only once. The quality of the learned mapping by \textcolor{blue}{U}-GENOT depends on the performance of the discrete solver. Therefore, as a representative example, we chose EfficientNet-B0 and extracted latent features from its different layers while varying the number of paired points to determine and tune an appropriate hyperparameter regime for \(\textcolor{blue}{\tau_{i}}\) and \(\epsilon\).

\begin{figure}[!h]
    \centering
    \includegraphics[width=.8\linewidth]{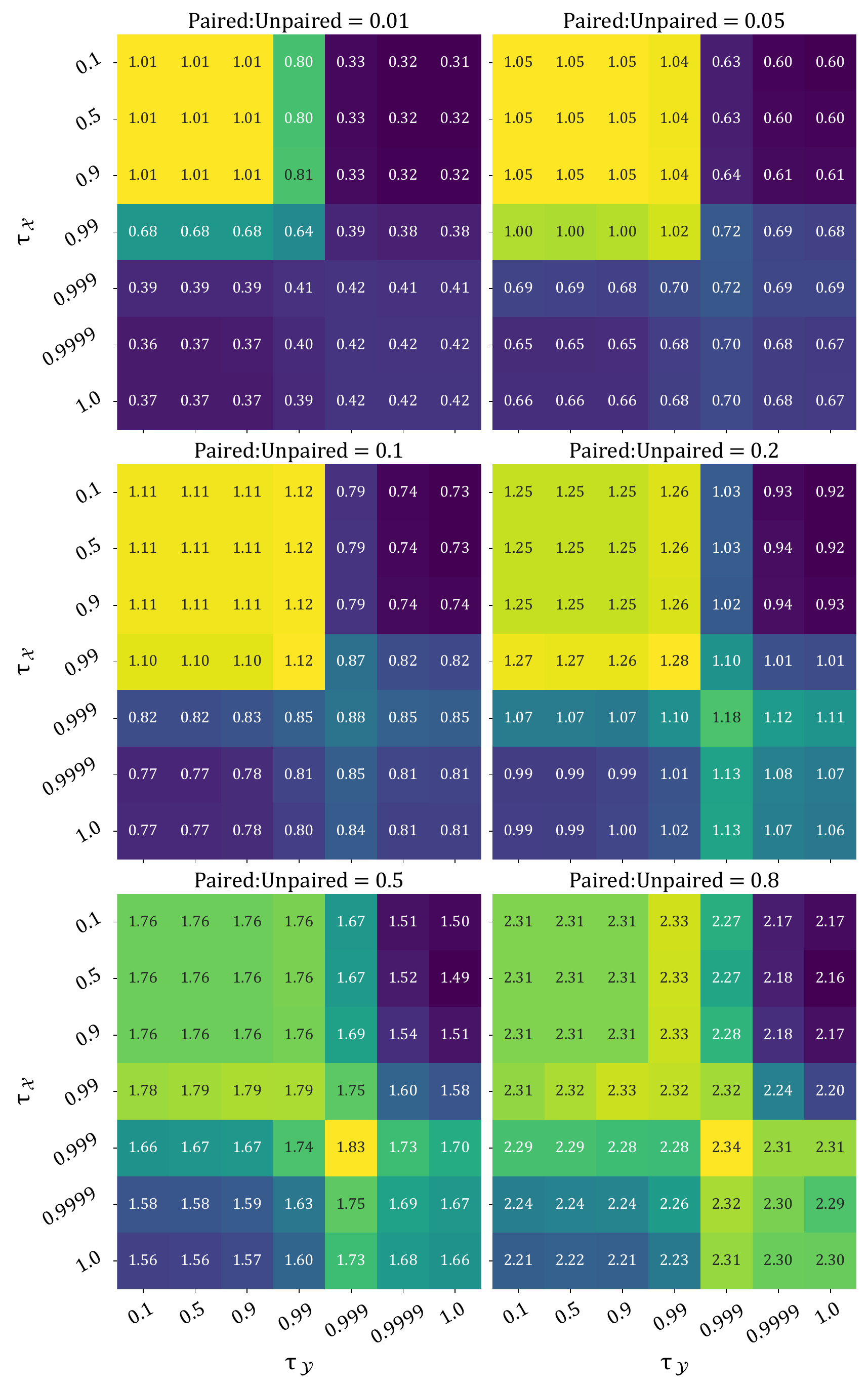}
    \caption{Heatmaps of the average values of the aes ratio as a function of \(\textcolor{blue}{\tau_\mathcal{X}}\) and \(\textcolor{blue}{\tau_\mathcal{Y}}\), with \(\epsilon = 10^{-3}\).}
    \label{fig:heatmaps_unbalanced}
\end{figure}

It is worth noting that the source and target distributions are uniform over the training set. However, when using the unbalanced setting, some samples may be excluded from the joint distribution by assigning them low probability. When the unbalancedness parameters are low (\(\textcolor{blue}{\tau_{i}} < 0.9\)), the true strategy is recovered, as the paired points have zero inter-space cost. However, this can lead to issues (Fig.~\ref{fig:heatmaps_unbalanced}) and requires tuning the hyperparameters to maximise matching accuracy under the optimal coupling \(\pi^\star_{\epsilon,\textcolor{blue}{\tau}}\), i.e., to minimise the number of excluded samples from the marginals using the validation set. We define the ratio of excluded samples for a coupling \(\pi\):

\begin{equation}
\text{excluded\_ratio}(\pi) = 1 - \frac{1}{N}\left| \left\{x_1,  \dots x_N \sim \pi_{\mathcal{X}} \right\} \right| -  \frac{1}{N}\left| \left\{y_1,  \dots y_N \sim \pi_\mathcal{Y} \right\} \right|~,
\end{equation}

where \(N\) is the total number of data points in the validation set and \(\left|\cdot\right|\) shows the set size. Using this, we defined the accuracy to excluded samples (i.e., aes) ratio to optimise this trade-off:  

\begin{equation}
 \text{aes} = \frac{\text{matching\_acc}(\pi)}{\text{excluded\_ratio}(\pi)}~.
\end{equation}

\begin{figure}[!t]
    \centering
    \includegraphics[width=0.4\linewidth]{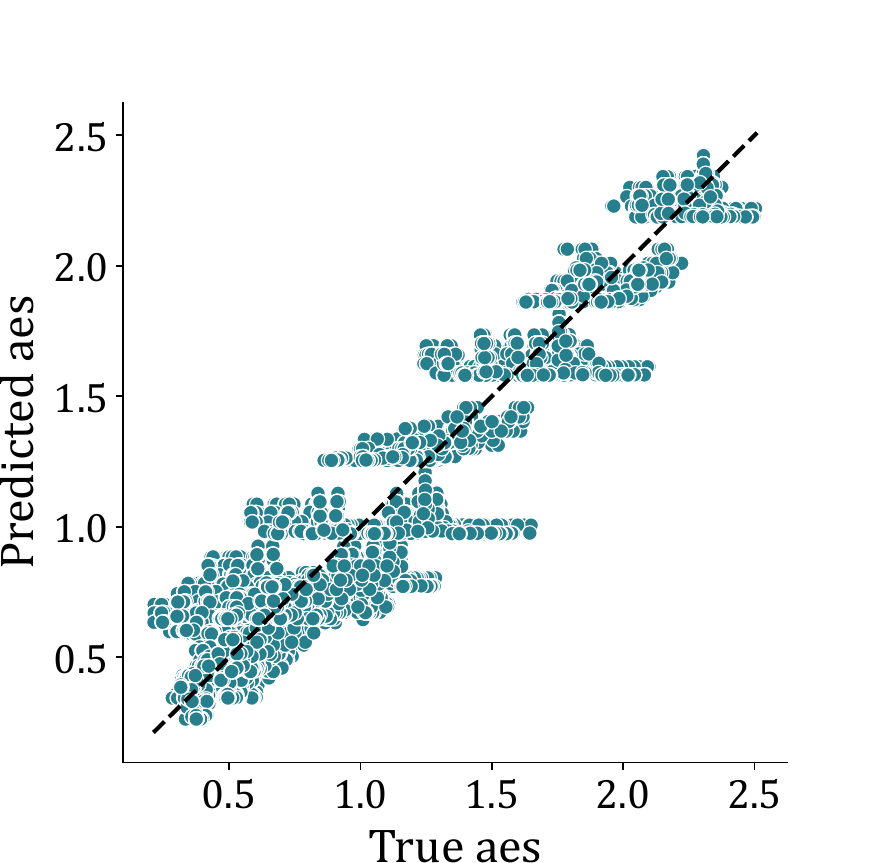}
    \caption{Predicted versus true aes ratio.}
    \label{fig:reg_ratio}
\end{figure}

To find a general rule for all layers and regions, we specified \( \epsilon \), \( \textcolor{blue}{\tau_\mathcal{X}} \), \( \textcolor{blue}{\tau_\mathcal{Y}} \), and the paired:unpaired ratio as independent variables, with aes ratio as the dependent variable. Fitting an ordinary least squares regression model resulted in an adjusted \( R^2 \) of 0.909 (Fig.~\ref{fig:reg_ratio}).

For the experiments reported in Sec.~\ref{sec:results::bio-art} we set the entropy regularisation parameter to \(\epsilon = 10^{-3}\) and show in Fig.~\ref{fig:heatmaps_unbalanced} the average value of the aes ratio for different pairwise combinations of \(\textcolor{blue}{\tau_\mathcal{X}}\) and \(\textcolor{blue}{\tau_\mathcal{Y}}\) as well as the paired:unpaired ratio. We found that \(\textcolor{blue}{\tau_\mathcal{X}} = \textcolor{blue}{\tau_\mathcal{Y}} = 0.99\) is suitable for varying levels of paired:unpaired sampled.

\newpage
\section{Additional results for biological-artificial neural representation alignment}

\begin{figure}[!h]
    \centering
    \includegraphics[width=.85\linewidth]{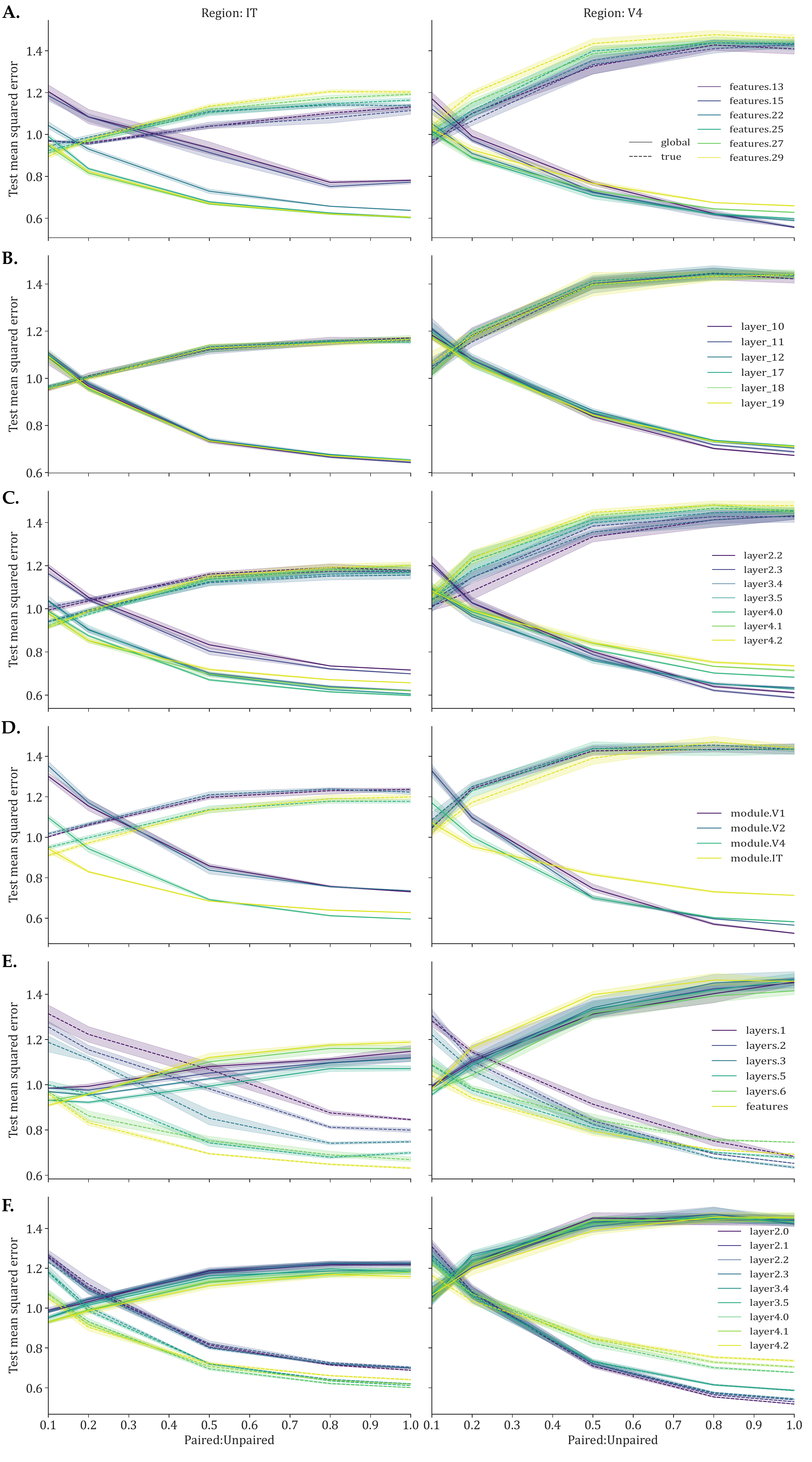}
    \caption{Test mean squared error of the learned conditional flow mappings from representations of different layers to IT (left) and V4 (right) for \textbf{A)} VGG16, \textbf{B)} ViT-L-32, \textbf{C)} ResNet50, \textbf{D)} CORNet-S, \textbf{E)} EfficientNet-B0, and \textbf{F)} Robust-ResNet50-L2-3.}
    \label{fig:majaj_pe_layers}
\end{figure}

\begin{figure}[!h]
    \centering
    \includegraphics[width=.85\linewidth]{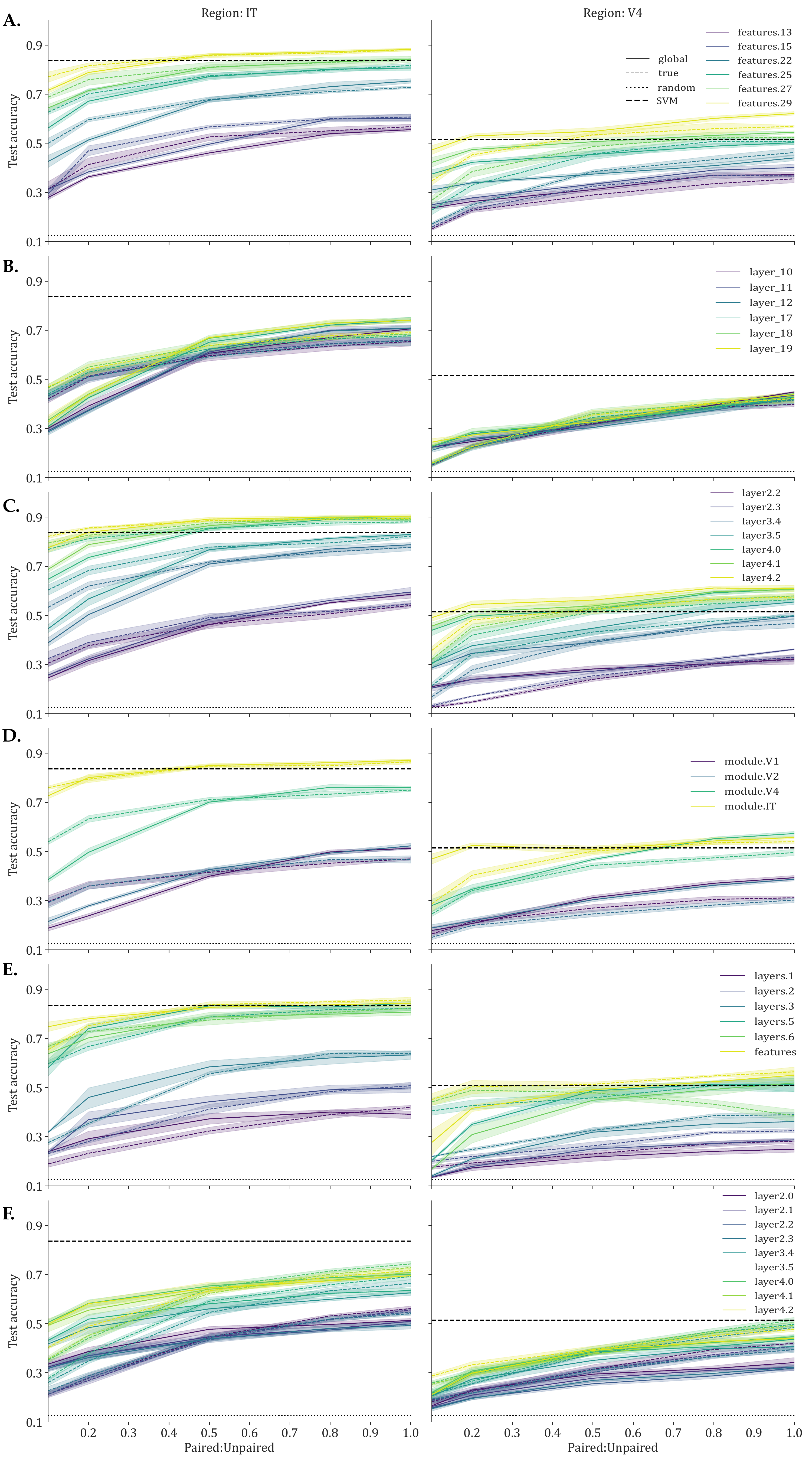}
    \caption{Downstream image category classification performance using learned conditional flow maps from different network layers, evaluated on IT (left) and V4 (right) for \textbf{A)} VGG16, \textbf{B)} ViT-L-32, \textbf{C)} ResNet50, \textbf{D)} CORNet-S, \textbf{E)} EfficientNet-B0, and \textbf{F)} Robust-ResNet50-L2-3.}
    \label{fig:majaj_acc_layers}
\end{figure}

\begin{figure}
    \centering
    \includegraphics[width=1\linewidth]{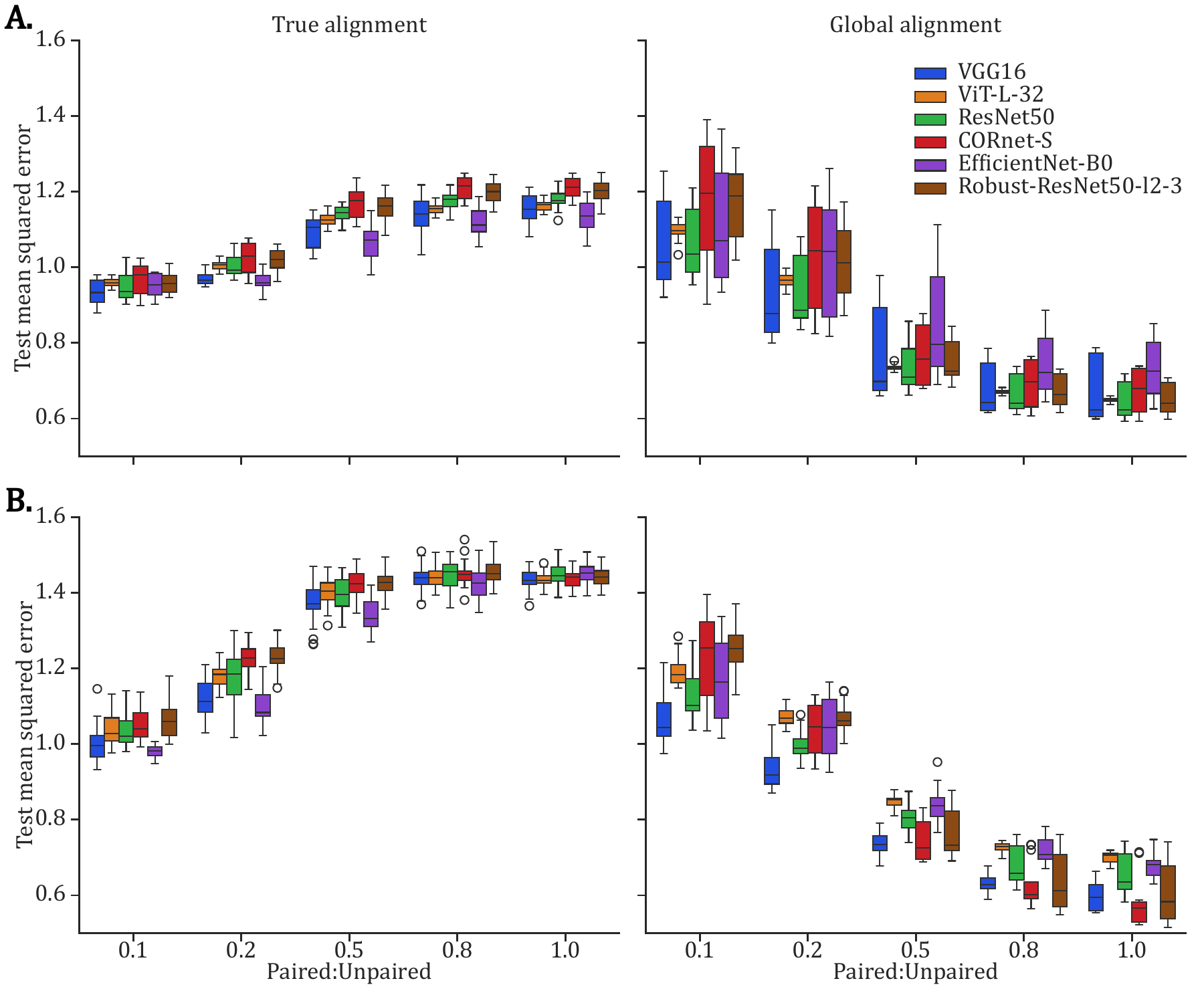}
    \caption{Boxplots of test mean squared error of the learned conditional flow matching from latent representations of all network layers to neural recordings in \textbf{A)} IT and \textbf{B)} V4, using true (left) and global (right) alignment across varying paired:unpaired ratios.}
    \label{fig:majaj_pe_box}
\end{figure}

\begin{figure}
    \centering
    \includegraphics[width=1\linewidth]{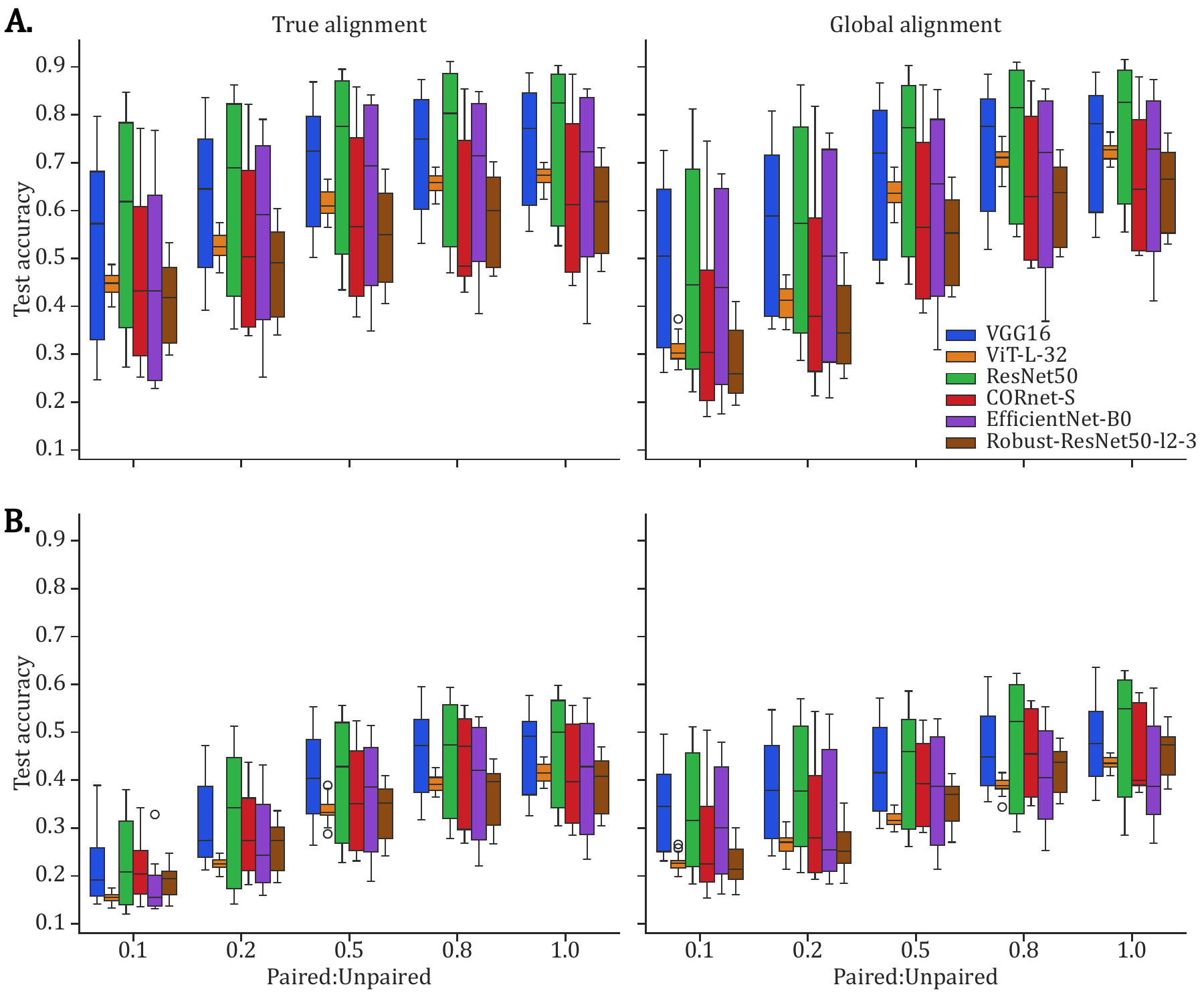}
    \caption{Boxplots of test accuracy of the learned conditional flow matching from latent representations of all network layers to neural recordings in \textbf{A)} IT and \textbf{B)} V4, using true (left) and global (right) alignment across varying paired:unpaired ratios.}
    \label{fig:majaj_acc_box}
\end{figure}

Here, we provide a breakdown of the results presented in Sec.~\ref{sec:results::bio-art}. Specifically, the test mean squared error of the learned conditional flow mappings (Fig.~\ref{fig:majaj_pe_layers}) for all the extracted layers of the selected models (Table~\ref{tab:arch-majaj}), and the corresponding downstream image category classification performance (Fig.~\ref{fig:majaj_acc_layers}). Fig.~\ref{fig:majaj_pe_box} and Fig.~\ref{fig:majaj_acc_box} present an aggregate of these results. 

\newpage

\end{document}